\pdfoutput=1

\documentclass[12pt,compress]{article}

\usepackage{float}
\usepackage{amssymb}
\usepackage{amsmath}
\usepackage{authblk}
\usepackage{graphicx}
\usepackage{multirow}
\usepackage{float}
\usepackage{tabularx}
\usepackage[margin=3cm]{geometry}
\usepackage{array}
\usepackage{booktabs}
\usepackage[table]{xcolor}
\usepackage{bm}
\usepackage{caption}
\usepackage{titlesec}
\titleformat{\section}
  {\normalfont\large\normalfont}
  {\thesection}{1em}{}

\titleformat{\subsection}
  {\normalfont\normalsize\normalfont}
  {\thesubsection}{1em}{}
  
\newcolumntype{L}[1]{>{\raggedright\arraybackslash}p{#1}}
\newcolumntype{Y}{>{\raggedright\arraybackslash}X}

\begin{document}

\begin{titlepage}
    {\raggedright
        {\textbf{Graphical Abstract}}
    }
    \begin{center}
    \includegraphics[width=1\textwidth]{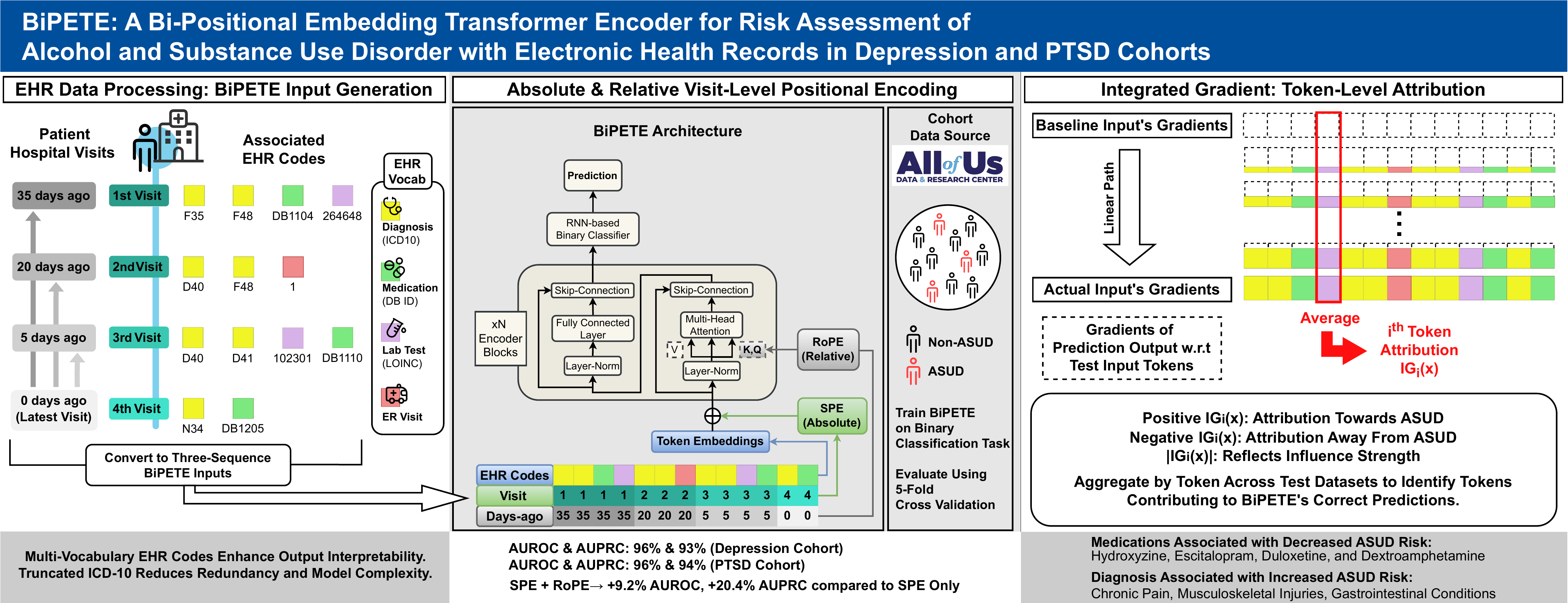}
    \vfill
    \end{center}
\end{titlepage}

\date{}
\title{\fontsize{19}{18}\selectfont BiPETE: A Bi-Positional Embedding Transformer Encoder for Risk Assessment of Alcohol and Substance Use Disorder with Electronic Health Records}

\author[1]{Daniel S. Lee}
\author[2]{Mayra S. Haedo-Cruz}
\author[1]{Chen Jiang}
\author[1]{Oshin Miranda}
\author[1]{LiRong Wang}

\affil[1]{\small{Department of Pharmaceutical Sciences, University of Pittsburgh School of Pharmacy, Pittsburgh, PA, 15213, USA}}
\affil[2]{\small{Division of Clinical and Translational Cancer Research, University of Puerto Rico Comprehensive Cancer Center, San Juan, PR, 00921, USA}}
\maketitle

\begin{abstract}
Transformer-based deep learning models have shown promise for disease risk prediction using electronic health records (EHRs), but modeling temporal dependencies remains a key challenge due to irregular visit intervals and lack of uniform structure. We propose a Bi-Positional Embedding Transformer Encoder or BiPETE for single-disease prediction, which integrates rotary positional embeddings to encode relative visit timing and sinusoidal embeddings to preserve visit order. Without relying on large-scale pretraining, BiPETE is trained on EHR data from two mental health cohorts–depressive disorder and post-traumatic stress disorder (PTSD)–to predict the risk of alcohol and substance use disorders (ASUD). BiPETE outperforms baseline models, improving the area under the precision-recall curve (AUPRC) by 34\% and 50\% in the depression and PTSD cohorts, respectively. An ablation study further confirms the effectiveness of the dual positional encoding strategy. We apply the Integrated Gradients method to interpret model predictions, identifying key clinical features associated with ASUD risk and protection, such as abnormal inflammatory, hematologic, and metabolic markers, as well as specific medications and comorbidities. Overall, these key clinical features identified by the attribution methods contribute to a deeper understanding of the risk assessment process and offer valuable clues for mitigating potential risks. In summary, our study presents a practical and interpretable framework for disease risk prediction using EHR data, which can achieve strong performance.
\end{abstract}



\clearpage
\section*{Introduction}
In recent decades, the growing availability of electronic health records (EHRs), together with advances in machine learning and deep learning, has opened new avenues for enhancing patient‐outcome prediction and have achieved notable successes \cite{rajpurkar22,hama25,rasmy22}. Hierarchical, longitudinal EHR data consists of large-scale, chronologically ordered records of patients’ diagnoses, treatments, laboratory tests, and health trajectories \cite{atasoy19}, and this structure presents significant opportunities for leveraging transformer-based deep learning models to predict outcomes and understand disease progression \cite{yangx22}. Despite these advances, a major challenge in predictive modeling for healthcare is capturing temporal dependencies and inter-visit correlations, as the disease progression, comorbidities, and treatment effects unfold over time \cite{rasmy21,devlin19}. EHR data are primarily organized by visits, and their irregular and sparse intervals further complicate fine-grained temporal modeling in EHR data \cite{xie22,holmes21}. To enable transformers to capture visit-level interactions among EHR tokens that reflect real-world clinical patterns, input EHR code sequences must be carefully modeled.

Transformer models have demonstrated efficacy in learning complex interdependencies within sequence data, a strength attributed to their attention mechanism \cite{yangx22,li20a,si21}. Consequently, several specialized transformer-based architectures have been introduced to model EHR data, each addressing temporal dependencies in distinct ways \cite{lindhagen15,valderas09}. BEHRT \cite{li20a} and Med-BERT \cite{rasmy21} both adopt the Bidirectional Encoder Representations from Transformers (BERT) \cite{devlin19} framework, pretraining on large-scale EHR data to learn contextual representation of the EHR tokens, with the goal of fine-tuning on a wide range of downstream tasks. To model the timing of EHR code occurrences, Med-BERT adds a learned positional embedding (LPE) applied to groups of codes within each patient visit, with an optional ordinal LPE assigned to individual codes \cite{rasmy21}. BEHRT adds two LPEs: age, and visit-segment embeddings. TransformEHR employs a transformer encoder-decoder structure to pretrain on the task of predicting ICD codes for future patient visits \cite{yangz23}. Similar to Med-BERT, it uses learned visit-level positional encoding, supplemented with sinusoidal positional embeddings (SPE) derived from visit dates to capture additional time information.

These systems demonstrate the effectiveness of transformers for modeling EHR data and a reliance on large-scale pretraining. However, both LPE and SPE are forms of absolute positional encoding (APE) and do not explicitly encode relative position information of the tokens. While APE specifies the position of each token by adding a unique positional vector to the token embeddings, the burden of learning relative dependencies is left to the subsequent attention and linear layers in the transformer block. Another limitation of adding multiple APEs is the introduction of noise. Redundant positional information could lead to signal superposition that distorts rather than enriches token representations. 

As in transformer-based natural language models, transformer-based models that input EHR sequence can benefit from encoding relative visit-level temporal information using RoPE. RoPE encodes positional information through position-dependent rotations of token embeddings. This rotation mechanism enables effective modeling of local relationships while reducing long-ranged dependency effects \cite{su24}. Leveraging these insights, we propose BiPETE, a Bi-Positional Embedding Transformer Encoder classifier that integrates both absolute and relative positional encodings–SPE and RoPE–which complement each other to capture temporal dynamics of longitudinal structured EHR data for disease prediction. BiPETE’s RoPE provides a positional encoding in which EHR token interactions are governed by their relative temporal separation in days between visits, enabling the model to leverage irregular, yet clinically meaningful time intervals. In contrast, BiPETE’s SPE encodes absolute visit order, reinforcing sequencing and grouping of the tokens.

To enhance the representation of heterogeneous EHR data, we integrate multiple standard clinical vocabularies. Unlike BEHRT, MedBERT, and TransformerEHR, BiPETE is not pretrained on large-scale EHR data, but trained on a moderate-sized dataset for a single-disease prediction task (Table1). While our approach reduces the data and computation requirements for deployment, the high-cardinality of multi-vocabulary medical codes can complicate token representation and pattern learning \cite{holmes21,rabbani22,bailly22}. To address this, we grouped diagnosis codes–which constitutes the largest proportion of our vocabulary–into broader categories that reflect common traits, related diseases, or conditions, thus preserving category-level information while reducing the vocabulary size.

We evaluate BiPETE on predicting the risk of subsequent alcohol and substance use disorders (ASUD) in two mental health disorder (MHD) cohorts derived from the National Institute of Health (NIH) All of Us (AoU) EHR data \cite{ramirez22}: depressive disorders and post-traumatic stress disorder (PTSD) cohorts. Balancing accuracy with explainability is another key challenge in applying deep learning to health-outcome prediction, as producing outputs that are interpretable and actionable for clinicians is essential \cite{sadeghi24}. To enhance clinical interpretability, we apply integrated gradients (IG), a gradient-based feature attribution method \cite{sundararajan17,abgrall24}, to aggregate token-level attributions, yielding global estimates of EHR tokens that drive BiPETE’s prediction for ASUD and non-ASUD classes.

\section*{Methods}

\section{Data Source and Cohort Definition}
We sourced our cohort data from the AoU Program, a NIH-funded cloud-based biomedical data repository and analysis platform. AoU data repository comprises clinical, genomic, demographic and other health-related data from de-identified participants aged 18 or older, collected across more than 50 healthcare organizations in the United States \cite{ramirez22}. We extracted standardized EHR data curated under Observational Medical Outcomes Partnership (OMOP) Common Data Model (CDM) from controlled-access tier. The analyses were conducted using data repository version 8, which includes EHR data from 394,596 participants. Data is available at AoU Research Hub upon registration.

Two cohorts of patients with MHD, depressive disorders and PTSD, were constructed using AoU Cohort Builder. AoU patient data are organized using standardized medical information referred to as “concepts”. The custom medical concept sets used to build our cohorts are reported in Supplementary Table 3 and 4. Prior to preprocessing, the depression and PTSD cohorts comprised 84,163 and 15,334 patients, respectively (Supplementary Table 1). Following data preprocessing, cohort sizes were reduced to 65,643 for depression and 9,310 for PTSD (Table 1).

\begin{table}[!t]
\small
\centering
\resizebox{155mm}{!}{
\renewcommand{\arraystretch}{1.2}
\noindent\begin{tabular}{
   >{\centering}m{35mm}
   >{\centering}p{60mm}
   >{\centering\arraybackslash}p{60mm}
}
\toprule
\multicolumn{1}{c}{\bfseries Cohort Stats} 
& \multicolumn{1}{c}{\bfseries Depressive Disorder} 
& \multicolumn{1}{c}{\bfseries PTSD} \\
\midrule
COHORT SIZE & 65,643 & 9,310\\
\midrule
CASES : CTRL & 20.6\% : 79.4\% & 24.7\% : 75.3\%\\
\midrule

\raisebox{-3\height}{VOCAB SIZE}
&
    \begin{tabular}[t]{
         >{\centering}p{1cm}
         >{\centering}p{0.8cm}
         >{\centering}p{1cm}
         >{\centering\arraybackslash}p{0.8cm}
      }
      \multicolumn{1}{c}{ICD10} 
        & \multicolumn{1}{c}{DB ID} 
        & \multicolumn{1}{c}{LOINC}
        & \multicolumn{1}{c}{ER} \\
      \midrule
      33.1\% & 34.6\% & 32.1\% & 0.2\%\\
      & & &\\[-2ex]
      \multicolumn{3}{p{3.6cm}}{\small{\% of N=4,904}}
    \end{tabular}
&
    \begin{tabular}[t]{
     >{\centering}p{1cm}
     >{\centering}p{0.8cm}
     >{\centering}p{1cm}
     >{\centering\arraybackslash}p{0.8cm} 
    }
    \multicolumn{1}{c}{ICD10} 
    & \multicolumn{1}{c}{DB ID} 
    & \multicolumn{1}{c}{LOINC}
    & \multicolumn{1}{c}{ER} \\
    \midrule
    41.6\% & 34.7\% & 23.4\% & 0.3\%\\
    & & &\\[-2ex]
    \multicolumn{3}{p{3.6cm}}{\small{\% of N=3,535}}
    \end{tabular} \\

\midrule

\raisebox{-1.2\height}{\shortstack{DATASET-WIDE\\FREQUENCY\\DISTRIBUTION}}
&
    \begin{tabular}[t]{
         >{\centering}p{1cm}
         >{\centering}p{0.8cm}
         >{\centering}p{1cm}
         >{\centering\arraybackslash}p{0.8cm}
      }
      \multicolumn{1}{c}{ICD10} 
        & \multicolumn{1}{c}{DB ID} 
        & \multicolumn{1}{c}{LOINC}
        & \multicolumn{1}{c}{ER} \\
      \midrule
      65.8\% & 22.1\% & 11.2\% & 0.9\%\\
      & & &\\[-2ex]
      \multicolumn{3}{p{3.6cm}}{\small{\% of N=4,712,501}}
    \end{tabular}
&
    \begin{tabular}[t]{
     >{\centering}p{1cm}
     >{\centering}p{0.8cm}
     >{\centering}p{1cm}
     >{\centering\arraybackslash}p{0.8cm} 
    }
    \multicolumn{1}{c}{ICD10} 
    & \multicolumn{1}{c}{DB ID} 
    & \multicolumn{1}{c}{LOINC}
    & \multicolumn{1}{c}{ER} \\
    \midrule
    65.5\% & 25.2\% & 8.3\% & 1.0\%\\
    & & &\\[-2ex]
    \multicolumn{3}{p{3.6cm}}{\small{\% of N=749,261}} \\
    \end{tabular} \\
\bottomrule
\end{tabular}
}
\caption{\textbf{Characteristics of Depression and PTSD Cohorts.} Cohort and data descriptions are generated after preprocessing EHR data into BiPETE input format. Vocabulary size denotes the size of EHR codes set for each standard vocabularies, whereas dataset-wide EHR code distribution refers to the frequency distribution of codes across the dataset. ICD10 statistics are derived after truncating codes to the first three characters.}
\label{table1}
\end{table}

For cohort EHR data preprocessing, we defined the following clinical scenario: a patient with MHD diagnosis—either depressive disorders or PTSD—but without a prior diagnosis of ASUD presents at a clinic. The current risk of new-onset ASUD is assessed based on the patient’s medical history from the preceding 15 months. The patient is labeled as a case if diagnosed with ASUD at a subsequent visit, and as a control if not diagnosed with ASUD (Figure 1). For control patients, the medical history from their most recent visits was used while ensuring no ASUD diagnosis was recorded during their entire observation period.

\begin{figure}[!t]
\centering
\includegraphics[width=1\textwidth]{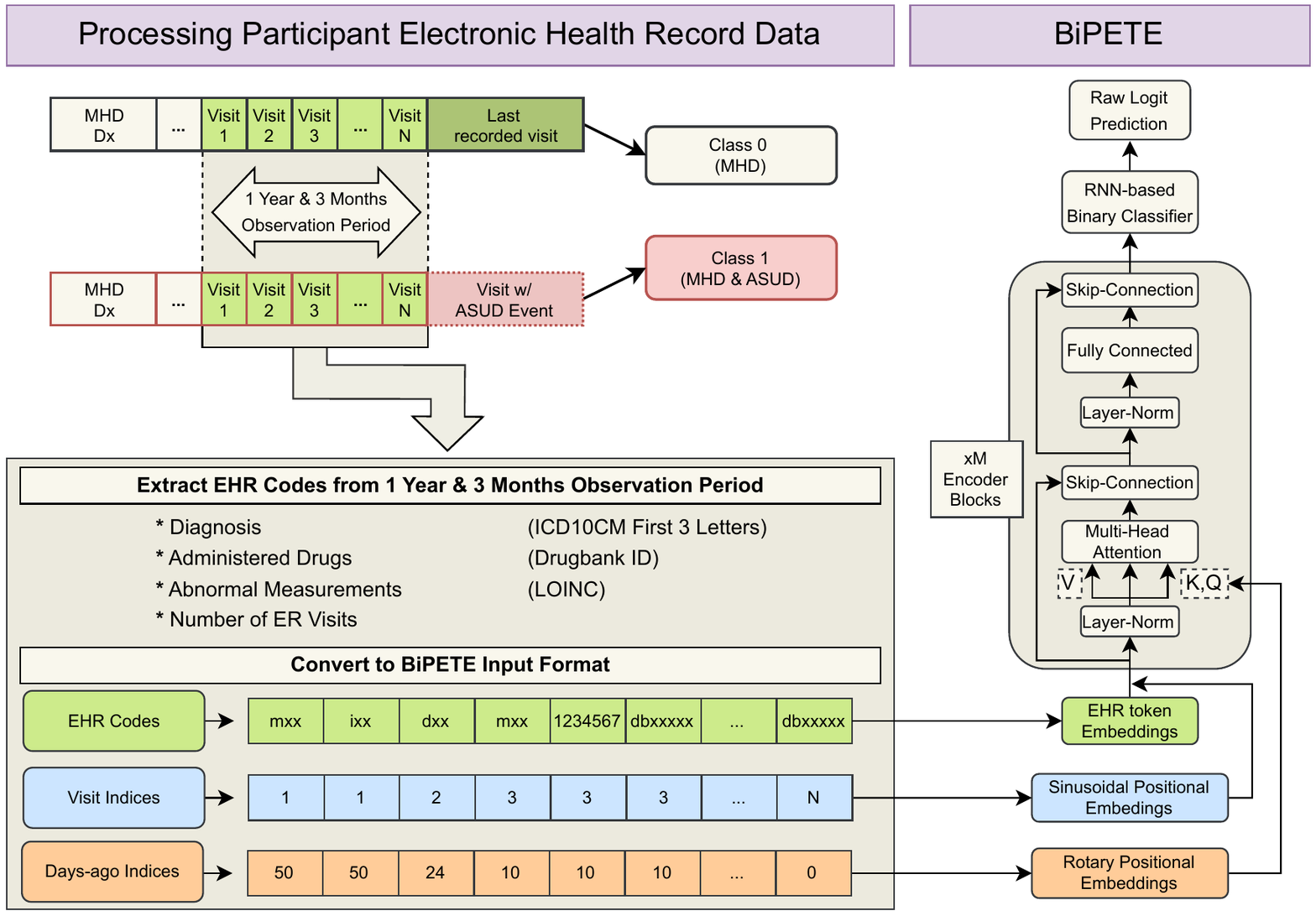}
\caption{\textbf{Pipeline Flowchart Illustrating Data Preprocessing, Model Input Construction, and BiPETE Architecture.} MHD Dx refers to diagnosis of Mental Health Disorder. For ASUD and non-ASUD classes, only the EHR codes from visits occurring after MHD diagnosis are extracted. Repeated values within the Visit and Days-ago index sequences indicate that the corresponding EHR codes were recorded during the same clinical visit. In the BiPETE architecture, the K, Q, V are the key, query and value embeddings, respectively, and M refers to the number of encoder blocks.}
\label{fig1}
\end{figure}

\section{EHR Preprocessing and Input Representation}
\subsection{Preprocessing Multi-Vocabulary EHR}
Structured EHRs offer a wealth of patient information encompassing diagnosis, medication, laboratory tests and other clinical data. To capture this rich clinical context, we constructed input instances using standard vocabularies, including the International Classification of Diseases, Tenth Revision (ICD10) for diagnosis, DrugBank Identifiers (DBID) for medications, and Logical Observation Identifiers Names and Codes (LOINC) for laboratory measurements. Laboratory tests were restricted to those flagged abnormal–either high or low–emphasizing clinically significant recordings. Additionally, the number of emergency room (ER) visits associated with each visit was incorporated as an indicator of acute events. 

OMOP CDM organizes clinical information using relational tables, requiring a series of preprocess steps to construct EHR input sequences. After defining our cohorts, we extracted medical concepts for diagnosis, medications and measurements for each patient using visit occurrence identifiers and their corresponding visit date ranges, defined by a start and end date. To reduce time-series fragmentation, patient visits occurring within a 7-day window were merged into a single visit, grouping temporally related encounters and analyzing them as a single episode of care. Patients with fewer than three visits within the observation window were excluded to ensure sufficient clinical context per input instance.

The extracted concepts were mapped to standardized vocabularies and consolidated to represent the clinical events occurring within visits to generate sequential EHR code data. There are visits with no recorded or extracted EHR codes, and a unique token was assigned to such visits to denote the absence of clinical information.

To reduce model complexity and sparsity in the input representation, ICD10 diagnosis codes were truncated to their three-character root categories, aggregating related diagnoses into broader groups. This grouping increased token frequency, mitigated redundancy and enhanced model efficiency by reducing vocabulary size. Furthermore, the patient's EHR history was restricted to a 15-month observation window to focus on information most predictive of our classification task.

\subsection{Temporal Modeling with Auxiliary Sequences}
BiPETE processes a sequence of EHR codes alongside two auxiliary sequences to encode temporality: visit indices and days-ago indices (Figure 1). In line with prior studies \cite{li20a,lindhagen15}, visit indices follow the chronological order of a patient’s visits, with the earliest visit indexed as 0. Days-ago indices are computed as the time difference in days between each visit and the most recent recorded visit in the patient's medical record, with codes from the most recent visit indexed as 0. The temporal information represented by visit indices and encoded using SPE is hereafter referred to as visit embeddings, while that represented by days-ago indices and encoded using RoPE is referred to as days-ago embeddings.

Days-ago embeddings capture the relational dynamics of EHR code occurrences between visits. RoPE applies rotation transformations to query and key embeddings of EHR tokens, such that the dot product between queries and keys depend on their relative positions, decaying interdependence with increased relative distance between them \cite{lindhagen15}. In simpler terms, RoPE explicitly expresses short- and long-ranged token relationships, assisting the transformer’s attention mechanism to learn the interaction patterns. All codes within a single visit undergo the same rotational transformation, effectively grouping them together into a visit while capturing relative temporal relationships of visits.

Patient visits are intermittent and visit gaps vary widely across patients. Applying RoPE directly to such varying days-ago indices can complicate model learning, as each index produces a different rotation based on the gap between visits. To strengthen visit order and token grouping, we encode visit order using SPE, applying the same absolute positional encoding to the codes recorded within a visit.

\section{BiPETE Architecture}
BiPETE is a transformer-based classifier adapted from BERT \cite{si21} with architectural modifications to the encoder and an addition of a bidirectional gated recurrent unit (BiGRU) classifier head (Figure 1). Unlike the original BERT, BiPETE employs the bidirectional self-attention encoder without leveraging pre-training. The model parameters are randomly initialized, and the token relationships are learned solely in the context of the single-disease classification task. Two modifications are applied to the encoder block: RoPE embeddings are applied to the query and key representations of tokens, and Pre-LayerNorm configuration is used. After the input EHR token sequence is processed by stacked encoder blocks, the GRU head generates a single output logit, which is passed through a sigmoid and optimized with Binary Cross-Entropy loss against the ASUD or non-ASUD label.

\section{Model Training and Evaluation}
\subsection{Model Training and Cross-validation}
We evaluated BiPETE with the ASUD risk prediction task on depressive disorders and PTSD cohorts and employed five-fold cross-validation to assess model generalizability \cite{rodriguez09}. For each iteration, the data was partitioned into training, validation and test subsets in a 7:1:2 ratio. Across the iterations, each fold served once as a test set. 

A 6-encoder-layer architecture was used to train and evaluate the model on the depression cohort, resulting in a 48-million-parameter model. For the PTSD cohort, a reduced 3-encoder-layer version was employed due to its smaller dataset size, resulting in a 35-million-parameter model. The size of large language models must scale with dataset size to improve performance. A study on compute-optimal training suggested that, for a given compute budget, model parameters and training data should be scaled at equal proportion, with approximately 20 training tokens per parameter to achieve optimal balance between performance and compute efficiency \cite{hoffmann22}. In our case, although the models were larger relative to the dataset size, ablation studies indicated that our configuration, combined with early stopping based on average validation loss across folds, yielded the best  performance. 

To assess the impact of the positional encodings on model performance, we conducted ablation experiments in the depression cohort by excluding visit or days-ago embedding, comparing training and test metrics for models configured with visit embedding alone, days-ago embedding alone, and both combined. For comparability, all models were trained using the same random seeds, deterministic computation settings and five-fold cross-validation protocol.

\subsection{Evaluation Metrics}
We report area under receiver operator characteristics curve (AUROC) and area under precision-recall curve (AUPRC) as primary performance evaluation metrics. Both AUROC and AUPRC deliver threshold-independent assessments by summarizing model performance across a range of decision thresholds. AUROC measures the trade-off between true-positive rate and false-positive rate and is relatively robust to class imbalance, as it can emphasize the model's ability to identify positives amidst a large proportion of negatives. AUPRC captures trade-off between precision and recall across all decision thresholds. It reflects the effect of missed cases via recall and is well-suited for evaluating disease risk prediction characterized by rare positive instances and high costs associated with missed detections. In addition to AUROC and AUPRC, we report positive predictive value (PPV), sensitivity, and specificity on three different thresholds to communicate interpretable evaluation of model performance. All metrics are reported as the mean and standard deviation across the five held-out test folds. Coefficient of variation (CV) across the thresholds is calculated across the models to compare the dispersion of the metrics.

\subsection{Baseline Models Used for Benchmarking}
We compared BiPETE against three baseline models: BiGRU, Logistic Regression (LR) and Bernoulli Naive Bayes (BNB). For LR and BNB, the EHR data was represented as one-hot encoded vectors, whereas for BiGRU, EHR codes were transformed to token embeddings without visit and days-ago embeddings. LR is a robust classifier and was included to assess whether the data exhibits linear separability. BNB was used to evaluate whether the presence or absence of EHR codes could predict ASUD risk, assuming conditional independence among the codes as in the Naive Bayes framework. BiGRU served as a deep learning baseline capable of modeling sequential dependencies in EHR data. All baseline models were trained and evaluated under the same five-fold cross-validation setup as BiPETE to ensure consistency.

\section{Attribution Analysis}
\subsection{Integrated Gradients}
To interpret the contribution of the EHR codes to ASUD prediction within our cohorts, we employed Integrated Gradient (IG) to compute token-level attributions. IG provides feature attributions that quantify the model’s sensitivity to changes in each input, capturing variations in each feature that affect the prediction \cite{sundararajan17}. Grounded in integral calculus, IG follows axiomatic principles that enable instance-level attributions to be naturally aggregated into global, dataset-level insights. 

IG requires a predefined, uninformative baseline input to serve as a reference for measuring attribution. The completeness axiom ensures that the sum of all feature attributions equals the difference between the model output and the reference output, while the sensitivity axiom guarantees that only features influencing the prediction receive nonzero attributions \cite{sundararajan17}. Tokens that consistently drive predictions away from the baseline yield positive or negative contributions, whereas uninformative features converge toward zero. In practice, IG computes the attribution for each token by integrating the gradient of the model output with respect to the input along a linear path from the baseline input to the actual input. Feature attribution $IG_i(x)$ of token i in input $x$ is formally written as: 
\begin{equation}
\operatorname{IG}_i(x)
= \bigl(x_i - x'_i\bigr)\cdot
\int_{\alpha=0}^{1}
\frac{\partial F\!\left(x' + \alpha\,(x-x')\right)}{\partial x_i}\, d\alpha
\label{eq1}
\end{equation}
where $x'$ denotes the baseline input vector, $\alpha$ $\in$ [0,1] is a scalar parameter tracing the straight-line path from $x'$ to $x$, and F represents the prediction function.
For deep learning models, this path integral is numerically approximated by averaging gradients over interpolated inputs. The approximation of $IG_i(x)$ of token i in input $x$ is formally written as:
\begin{equation}
\operatorname{IG}_i(x)
\approx (x_i - x'_i)\cdot
\frac{1}{m}\sum_{k=1}^{m}
\frac{\partial F\!\left(x' + \frac{k}{m}\,(x - x')\right)}{\partial x_i}
\label{eq2}
\end{equation}
where $x'$ denotes the baseline input vector, $x$ is the input being explained, $F$ represents the prediction function, $m$ is the number of interpolated steps and $k$ is the index of the steps. In our study, the baseline input is defined as a sequence of padding tokens. To ensure consistency in the attribution of tokens across instances with varying sequence length and positions, we retain the visit and days-ago embeddings of the original input when constructing the baseline.

IG token attributions quantify the extent to which each feature influences the predicted label, with positive values increasing and negative values decreasing the predicted likelihood. In our ASUD classification task, the attributions of EHR codes reflect the model’s internal reasoning rather than absolute risk.

\subsection{Aggregating Token Contribution: Relative Contribution}
We compute the Relative Contribution (RC) to quantify the directional attribution at the token level, defined as the ratio of a token’s average attribution in the true positive (TP) group to its average attribution in the true negative (TN) group. Only instances correctly predicted by the model are used in the RC calculation to extract meaningful attribution interpretation \cite{li20a}. Tokens showing opposite attribution signs across TP and TN groups were excluded, as they indicate inconsistent directional effects. Denoting $A_{TP}(t)$ as the average attribution of token $t$ in the true positive (TP) group, and $A_{TN}(t)$ as the average attribution in the true negative (TN) group, the RC of token $t$ is then calculated as: 
\begin{equation}
RC(t) = \frac{A_{\mathrm{TP}}(t)}{A_{\mathrm{TN}}(t)}
\label{eq3}
\end{equation}
RC values greater than 1 indicate contribution toward the ASUD class, whereas values less than 1 indicate contribution toward the non-ASUD class. To ensure statistical robustness and reduce the influence of outliers, we restrict our analysis on tokens that appear in at least 1\% of instances within each TP and TN group. For tokens occurring multiple times within a single instance, attribution values are averaged. RC values are calculated using the attribution values from the five held-out test sets.

\section*{Results}
\addcontentsline{toc}{section}{Results}
\setcounter{section}{0}
\section{Cohort Characteristics After Preprocessing}
Cohort EHR vocabulary sizes and distributions are shown in Table 1. Within the depression cohort, ICD10, DBID, and LOINC codes contribute approximately equally to the total EHR vocabulary, whereas in the PTSD cohort, the distribution is more uneven. Nevertheless, the dataset-wide frequency distributions are similar across both cohorts. Truncating ICD10 codes reduced model complexity and redundancy in clinical information. While this introduces loss in diagnostic specificity, it reduced the ICD10 vocabulary size by 95\%, from 32,662 unique codes to 1,625 in the depression cohort and by 93\%, from 21,137 unique codes to 1,469 in the PTSD cohort. Most data loss occurred during the extraction and mapping of visit-level EHR codes to standardized vocabularies–ICD10, DBID, and LOINC–which reduced the number of usable visits in many patient records. As a result, 22\% of samples in the depression cohort and 39\% in the PTSD cohort were removed (Supplementary Table 1).

\section{BiPETE Performance on ASUD Risk Prediction}
AUROC and AUPRC scores of BiPETE's ASUD risk prediction in depression disorder and PTSD cohorts are shown in Table 2.  On the AUROC, BiPETE with both days-ago and visit embeddings obtain the best results in the depression cohort with a score of 96.46\%, outperforming the baseline models–BiGRU, LR, and BNB–which scored 85.21\%, 83.54\% and 77.17\%, respectively. The AUPRC score of 93.18\% shows a performance gain of over 30\% compared to the three baseline models. A similar pattern for both metrics is observed in the PTSD cohort.

Supplementary Table 3 compares PPV, sensitivity and specificity of BiPETE, BiGRU and LR in the decision thresholds of 0.2, 0.5 and 0.8. As expected, increasing the decision threshold leads to higher PPV and specificity, but lower sensitivity for all models. Though this trend is consistent, the CV values for BiPETE metrics remain consistently lower than those of BiGRU and LR, indicating BiPETE achieves more stable and robust class separation across thresholds.

\begin{table}[!ht] 
\small
\centering
\resizebox{125mm}{!}{
\renewcommand{\arraystretch}{1.4}
\begin{tabular}{llll}
\toprule
& \textbf{Models} & \textbf{AUROC (\%)} & \textbf{AUPRC (\%)} \\
\hline
\multirow{6}{*}{\shortstack{\textbf{Depressive} \\ \textbf{Disorder}}}
& BiPETE & $96.46 \; (\pm 0.21)$ & $93.18 \; (\pm 0.27)$ \\
& RoPE Only& $94.46 \; (\pm 0.88 )$& $90.73 \; (\pm 1.17)$\\
& SPE Only& $88.33 \; (\pm 0.35)$& $77.34 \; (\pm 0.48)$\\
& BiGRU & $85.21 \; (\pm 0.24)$& $69.42 \; (\pm 0.35)$\\
& Linear Regression & $83.54 \; (\pm 0.47)$& $67.79 \; (\pm 0.30)$\\
& Bernoulli Naive Bayes & $77.17 \; (\pm 0.51)$& $46.73 \; (\pm 0.77)$\\
\hline
\multirow{4}{*}{\textbf{PTSD}} 
& BiPETE & $96.50 \; (\pm 0.40)$ & $94.04 \; (\pm 0.63)$ \\
& BiGRU & $80.13 \; (\pm 0.53)$& $62.29 \; (\pm 0.24)$\\
& Linear Regression & $83.97 \; (\pm 0.68)$& $70.86 \; (\pm 1.53)$\\
& Bernoulli Naive Bayes & $93.59 \; (\pm 5.44)$& $87.42 \; (\pm 9.33)$\\
\bottomrule
\end{tabular}
}
\caption{\textbf{AUROC and AUPRC Scores for ASUD Risk Prediction using BiPETE and Baseline Models in Depression and PTSD Cohorts.} The depression cohort includes two additional rows for positional encoding variants: one using only days-ago embeddings (RoPE) and the other only visit embeddings (SPE). The baseline BiGRU shares the same configuration as the BiGRU used as the classifier head in BiPETE. ±indicates standard deviation across five-fold cross-validation.}
\label{table2}
\end{table}

\section{Impact of Visit and Days-ago Embeddings on Performance}
In the depression cohort, we report AUROC and AUPRC scores of BiPETE and its variants configured with only days-ago embeddings or only visit embeddings to compare the impact of different positional embedding strategies. BiPETE outperforms the single positional embedding models in both metrics (Table 2). Compared to BiGRU, BiPETE, days-ago-embedding-only model, and visit-embedding-only model improved in AUPRC by 34.23\%, 30.70\% and 11.41\%, respectively. In addition to achieving higher AUROC and AUPRC scores, BiPETE exhibits lower standard deviations than the single-positional-embedding models, indicating a more consistent performance than the models with a single positional encoding. The average ROC and PR curves of the cross-validation folds for the BiPETE and its positional encoding variant models are presented in Supplementary Figure 1.

Figure 2 illustrates the learning behavior of the positional encoding variant models during training. We report training and validation metrics–including loss, accuracy and ROAUC–recorded over the course of 30 epochs. Among the configurations, BiPETE demonstrates the fastest convergence and exhibits the most stable training performance, indicated by the narrow error bars across all training metrics. BiPETE’s train accuracy reaches close to 100\%, showing near-perfect fit to the training data and signs of overfitting, as evidenced by the increase in validation loss after epoch 15. The days-ago-embedding-only model achieves validation metrics comparable to those of BiPETE, but the validation loss and accuracy exhibit high instability during training, reflected in the large standard deviations. The visit-embedding-only model shows training stability similar to the model with both embeddings, but underperforms across both training and validation metrics.

Supplementary Figure 2 presents the attention matrices of the nine heads in the final BiPETE layer for a representative instance, illustrating the effect of visit and days-ago embeddings. Each square matrix shows token-to-token attention scores of the input sequence. Both axes are annotated with the EHR vocabulary type and the visit indices, with the y axis representing tokens attending to those on the x axis. Ideally, different heads capture distinct aspects of the input EHR sequence, enabling the model to extract meaningful patterns for prediction, and the visualization shows that individual heads attend to different visit pairs. Tokens tend to exhibit high or low attention in groups according to their visit indices. Attention scores vary within the visits as well. Together, inter- and intra-visit dynamics can be seen in the attention maps.

\begin{figure}[!t]
  \centering
  \includegraphics[width=1\textwidth]{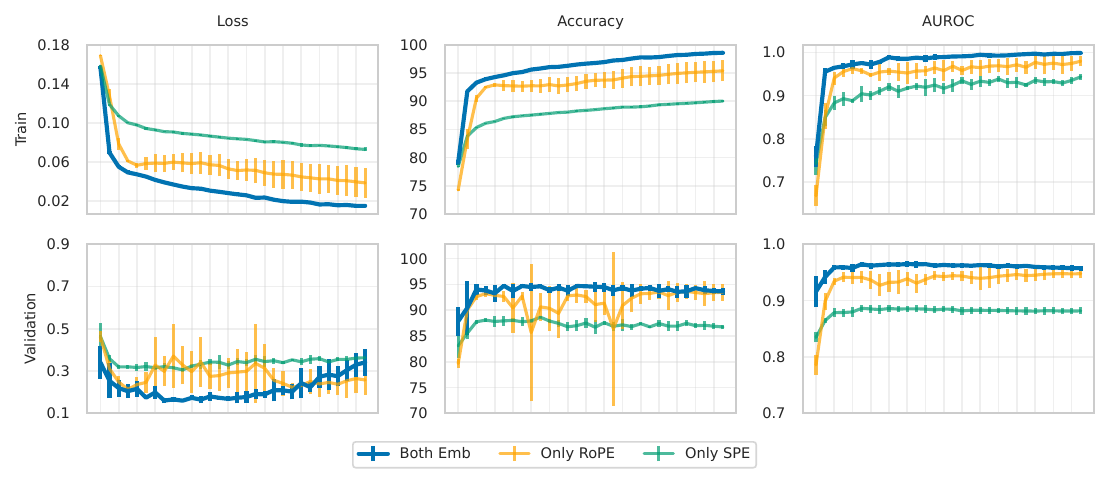}
  \caption{\textbf{Training and Validation Performance of Classifiers with Different Positional Encoding Configurations.} Training and validation loss, accuracy and AUROC are reported across 30 training epochs to compare model learning and generalization. The error bars indicate the standard deviation across five-fold cross-validation folds.}
  \label{fig2}
\end{figure}

\section{IG-Derived Indicators of Increased and Reduced ASUD Risk}
We performed token-level feature attribution using IG and calculated RC by averaging the token attributions of correctly predicted test samples. RC quantifies the influence of each token on BiPETE's prediction in the true positive group relative to the true negative group. We identified potential markers that drove BiPETE's prediction toward ASUD or Non-ASUD in both the Depression and PTSD cohorts. The identified indicators–including, abnormal lab test results, medication use, and comorbidities–are reported for each cohort.

\subsection{Indicators Associated with Higher ASUD Risk in Depression Cohort–Table 3}
Alterations in lymphocyte-related markers, including the neutrophil-to-lymphocyte ratio (NLR) and platelet-to-lymphocyte ratio reflect chronic inflammation that can influence neurotransmission and brain function, with evidence suggesting both neurotoxic and neuroprotective effects depending on context \cite{karatoprak21,gasparyan19}. Depression and SUD frequently co-occur and share overlapping immune dysfunction, as substances like alcohol and opioids directly impair immune function, while elevated pro-inflammatory cytokines correlate with depressive symptoms. Similarly, coagulation markers, such as prolonged prothrombin time (PT) and altered international normalized ratio (INR), are affected by substance use particularly chronic alcohol use through liver dysfunction, and by depression via medication adherence challenges, highlighting the bidirectional interaction between immune and hemostatic pathways and psychiatric and substance-related outcomes \cite{gasparyan19}. 

\begin{table}[!b]
\centering
\scriptsize 
\resizebox{155mm}{!}{
\renewcommand{\arraystretch}{1.2}
\begin{tabularx}{\textwidth}{
  >{\raggedright\arraybackslash}p{6.6cm}
  >{\raggedleft\arraybackslash}p{3.4cm}
  >{\raggedleft\arraybackslash}p{1.5cm}
  >{\centering\arraybackslash}p{2.4cm}
}
\toprule
\textbf{Indicators of High Risk} & \textbf{Risk Marker Type} & \textbf{RC} & \textbf{N (Case/Ctrl)} \\
\midrule
Lymphocytes / Total Leukocytes in Blood & Abnormal Lab test result & 50.7491 & 912/2703 \\
International Norm Ratio (Platelet Poor Plasma) & Abnormal Lab test result & 22.8152 & 408/938\\
Prothrombin Time (PT) & Abnormal Lab test result & 11.1948 & 560/1262 \\
\midrule
Vancomycin & Medication use & 5.2091 & 262/982 \\
Metronidazole & Medication use & 2.4416 & 522/1515\\
Acyclovir & Medication use & 2.2308 & 156/520 \\
\midrule
Other Peripheral Vascular Diseases & Comorbidities & 40.5746 & 358/1575 \\
Other Disorders of Thyroid & Comorbidities & 21.1796 & 175/717\\
Flatulence and Related Conditions & Comorbidities & 0.0644 & 300/1317 \\
Hereditary and Idiopathic Neuropathy & Comorbidities & 15.3311 & 360/811\\
Essential (primary) Hypertension & Comorbidities & 9.7897 & 4974/20857 \\
\bottomrule
\end{tabularx}
}
\caption{\textbf{Key Indicators Associated with Increased Risk of ASUD in Patients with Depressive Disorder.}}
\label{table3}
\end{table}
Vancomycin, while not psychoactive, may intersect with ASUD risk in patients with recent substance use or complex medical conditions, as altered pharmacokinetics and hospitalization-related stressors can create vulnerabilities for maladaptive substance behaviors \cite{sharma19}. Metronidazole may increase susceptibility to ASUD via disulfiram-like reactions when alcohol is consumed during or shortly after therapy, producing aversive physiological effects such as nausea, tachycardia, and abdominal discomfort that can promote maladaptive drinking or relapse \cite{alonzo19}. Acyclovir carries potential for neurotoxic side effects, particularly in older adults, patients with renal impairment, or those with multiple comorbidities, with symptoms such as confusion, ataxia, and altered mental status linked to elevated drug or metabolite concentrations and blood–brain barrier disruption \cite{martinez11}. These neurocognitive effects may impair mood regulation and inhibitory control, creating conditions for self-medication with alcohol or other substances.  

Conditions such as peripheral vascular disease, thyroid disorders, flatulence and related gastrointestinal conditions, hereditary neuropathies, and primary hypertension often require ongoing management and can involve chronic pain, fatigue, or functional impairment. These stressors may dysregulate the hypothalamic-pituitary-adrenal axis and alter dopaminergic and serotonergic signaling in reward pathways, increasing susceptibility to substance misuse as a coping strategy. Additionally, chronic illness and comorbidities can contribute to systemic inflammation, oxidative stress, and neuroplasticity alterations, which may amplify depressive symptoms and further predispose patients to SUD. The cumulative physiological burden, along with potential medication side effects and impaired autonomic or gastrointestinal function, can create a cycle of heightened stress and vulnerability, highlighting the importance of comprehensive medical and psychiatric management to mitigate substance misuse risk in these populations.

\subsection{Indicators Associated with Lower ASUD Risk in Depression Cohort–Table 4}
Platelets and red blood cell indices, such as mean corpuscular hemoglobin concentration (MCHC), may play interconnected roles in depression and ASUD through mechanisms involving serotonin signaling, inflammation, and oxygen delivery \cite{ng19,li24}. Platelets, which store the majority of the body’s serotonin, reflect central serotonergic activity and may be hyperactivated in depression \cite{martin23}. Chronic use of substances such as alcohol, cocaine, heroin, and methamphetamine can induce thrombocytopenia via bone marrow suppression, immune-mediated destruction, or increased consumption, further impairing neurocognitive function and contributing to mood dysregulation. Similarly, low MCHC, often resulting from iron deficiency anemia or chronic substance use, reduces oxygen delivery to neurons, disrupting neurotransmitter synthesis and exacerbating depressive symptoms \cite{ng19,li24}. Platelet abnormalities and low MCHC typically reflect the physiological consequences of substance use and depression, highlighting the potential benefit of interventions targeting these parameters to support mood stabilization and reduce indirect risk factors for SUD. 

\begin{table}[!t]
\centering
\scriptsize 
\resizebox{155mm}{!}{
\renewcommand{\arraystretch}{1.2}
\begin{tabularx}{\textwidth}{
  >{\raggedright\arraybackslash}p{6.6cm}
  >{\raggedleft\arraybackslash}p{3.4cm}
  >{\raggedleft\arraybackslash}p{1.5cm}
  >{\centering\arraybackslash}p{2.4cm}
}
\toprule
\textbf{Indicators of Low Risk} & \textbf{Type of Risk Marker} & \textbf{RC} & \textbf{N (Case/Ctrl)} \\
\midrule
MCHC [Mass/volume] & Abnormal Lab test result & 0.4432 & 1152/4102 \\
Platelets [\#/volume] in Blood & Abnormal Lab test result & 0.9242 & 290/606\\
\midrule
Naloxone & Medication use & 0.0115 & 253/928 \\
Cefazolin & Medication use & 0.0236 & 418/2721 \\
Amoxicillin & Medication use & 0.0457 & 867/4141 \\
Famotidine & Medication use & 0.0483 & 498/3312 \\
Dexamethasone & Medication use & 0.3031 & 521/2962 \\
Dextroamphetamine & Medication use & 0.3056 & 169/903 \\
Clavulanic acid & Medication use & 0.3480 & 440/2260 \\
Duloxetine & Medication use & 0.3664 & 648/2621 \\
\midrule
Other Specified Health Status & Comorbidities & 0.0009 & 403/2811 \\
Intracranial Injury & Comorbidities & 0.0031 & 216/680 \\
Other Extrapyramidal and Movement Disorders & Comorbidities & 0.0067 & 536/2429 \\
Male Erectile Dysfunction & Comorbidities & 0.0095 & 368/1168 \\
\bottomrule
\end{tabularx}
}
\caption{\textbf{Key Indicators Associated with Decreased Risk of ASUD in Patients with Depressive Disorder.}}
\label{table4}
\end{table}

Naloxone, a rapid opioid antagonist, remains a cornerstone of harm reduction, with community-based distribution programs achieving survival rates above 92–98\% \cite{bohler23,bazazi10,fischer25,petrovitch24}. $\beta$-lactam antibiotics such as cefazolin \cite{alasmari16,rao15,weiland15}, amoxicillin \cite{mergenhagen20,hakami16}, and clavulanic acid \cite{callans24,philogene22,maser24,schroeder14} modulate glutamatergic signaling by upregulating GLT-1 and the cystine-glutamate exchanger in addiction-relevant brain regions, attenuating ethanol or cocaine reward, reducing cue-induced reinstatement, and facilitating extinction of drug-seeking behaviors. Other agents, including hydroxyzine (especially in combination with 5-HT3 antagonists), famotidine \cite{mather20}, and dexamethasone \cite{aouizerate06,capasso97}, show promise in mitigating withdrawal symptoms, stress, or neuroinflammation, which are key risk factors for relapse. Psychostimulants like dextroamphetamine \cite{aouizerate06,smith24,chang14,mariani07}, when administered under clinical supervision, reduce substance use risk by addressing underlying neurochemical and behavioral deficits, supported by both population-level and preclinical data. 

Conditions such as other specified health status, intracranial injury, extrapyramidal and movement disorders, and male erectile dysfunction are generally chronic but manageable, often requiring structured medical oversight or ongoing outpatient care \cite{musco19}. These conditions may promote regular engagement with healthcare providers, facilitate adherence to treatment routines, and provide early opportunities for monitoring mental health, all of which can reduce reliance on maladaptive coping strategies such as substance use. While these disorders may impact daily functioning, their generally predictable course allows patients to maintain relative emotional and physiological stability, limiting the severity of depressive symptoms that often drive self-medication. 

\subsection{Indicators Associated with Higher ASUD Risk in PTSD Cohort–Table 5}
We found the presence of abnormalities in hematological and metabolic markers which may highlight stress-related dysregulation in PTSD and comorbid substance use disorder (SUD). Serum albumin and albuminuria reflect hepatic and renal reserve, with hypoalbuminemia and elevated albuminuria signaling systemic inflammation and vascular risk \cite{annoni90}. Mean platelet volume reflects systemic stress and shows mixed patterns in PTSD, with elevated levels indicating platelet activation and inflammation linked to SUD \cite{martin23}. Mean corpuscular hemoglobin abnormalities, often worsened by alcohol use, contribute to fatigue and cognitive impairment.  Together, these markers highlight the compounded biological burden of PTSD and SUD.

Several commonly prescribed medications can influence the risk of SUD in patients with PTSD, with effects varying based on neurobiological vulnerability and prior substance use history. Hydrocodone, widely used for pain management, is strongly associated with increased risk of opioid misuse and dependence in trauma-exposed adults, particularly among women, younger adults (18–34 years), and those with prior SUD, likely due to dysregulated dopaminergic and endogenous opioid pathways that amplify its reinforcing euphoric effects \cite{darwish17,miller04,schwartz06}. Oxybutynin has also been reported in case series to carry abuse potential because of hallucinogenic and euphoric CNS effects, suggesting heightened vulnerability in patients with prior SUD \cite{gulsun06,welk22}. Trimethoprim–sulfamethoxazole (TMP-SMX) has been implicated in neuropsychiatric adverse effects such as hallucinations, particularly in older adults, which may indirectly elevate SUD risk \cite{iqbal22,stuhec14}. Aripiprazole exhibits mixed effects: long-acting injectable formulations have been associated with improvements in psychiatric symptoms and reductions in alcohol and cocaine use among patients with PTSD or schizophrenia and co-occurring SUD, while other reports caution that it may exacerbate compulsive substance use urges in some individuals, consistent with FDA warnings on impulse control disorders. 

Neurological conditions such as mononeuropathies of the lower limb and musculoskeletal injuries (dislocations and sprains of the ankle, foot, and toe) contribute to chronic pain and functional impairment, often leading to self-medication. Dental caries and fibroblastic disorders reflect ongoing disease burden that may exacerbate stress and promote maladaptive coping. Alongside psychiatric comorbidities and high healthcare utilization, these conditions collectively amplify the biological and psychological burden seen in these patients \cite{scholz24,pastore18,berlt25}.

\begin{table}[!t]
\centering
\scriptsize 
\resizebox{155mm}{!}{
\renewcommand{\arraystretch}{1.2}
\begin{tabularx}{\textwidth}{
  >{\raggedright\arraybackslash}p{6.6cm}
  >{\raggedleft\arraybackslash}p{3.4cm}
  >{\raggedleft\arraybackslash}p{1.5cm}
  >{\centering\arraybackslash}p{2.4cm}
}
\toprule
\textbf{Indicators of High Risk} & \textbf{Risk Marker Type} & \textbf{RC} & \textbf{N (Case/Ctrl)} \\
\midrule
Albumin [Mass/Volume] in Serum/Plasma & Abnormal Lab test result & 80.5480 & 133/231 \\
Platelet mean volume [Entitic Volume] in Blood & Abnormal Lab test result & 38.7960 & 40/154 \\
Mean corpuscular hemoglobin [Entitic Mass] & Abnormal Lab test result & 3.6885 & 171/358 \\
\midrule
Hydrocodone & Medication use & 138.1751 & 192/364 \\
Oxybutynin & Medication use & 30.3633 & 35/97 \\
Trimethoprim/Sulfamethoxazole & Medication use & 16.4505 & 69/182 \\
Aripiprazole & Medication use & 10.1848 & 85/215 \\
\midrule
Mononeuropathies of Lower Limb & Comorbidities & 270.3382 & 50/146 \\
Dislocation/Sprain of Ankle, Foot, or Toe & Comorbidities & 26.4504 & 60/109 \\
Dental Caries & Comorbidities & 20.9523 & 169/345 \\
Fibroblastic Disorders & Comorbidities & 19.0740 & 73/214 \\
\bottomrule
\end{tabularx}
}
\caption{\textbf{Key Indicators Associated with Increased Risk of ASUD in Patients with PTSD.}}
\label{table5}
\end{table}

\subsection{Indicators Associated with Lower ASUD Risk in PTSD Cohort–Table 6}
Iron, particularly ferritin, and Vitamin B12, which are critical micronutrients, were found to influence the risk of ASUD in patients with PTSD. Vitamin B12 deficiency has been linked to both PTSD and SUD, likely due to its essential role in monoamine neurotransmitter synthesis and mood regulation. Lower B12 levels in individuals with alcohol or methamphetamine use have been associated with increased relapse risk, and supplementation may help restore energy, cognitive function, and emotional regulation, indirectly mitigating ASUD risk by reducing self-medication behaviors. Similarly, iron homeostasis, especially ferritin levels, is vital for dopaminergic neurotransmission, oxidative stress regulation, and myelination. Disruption of iron balance can lead to neurocognitive impairments and behavioral vulnerabilities relevant to PTSD and SUD, with animal studies showing that iron deficiency alters ferritin, dopamine metabolism, and neuroproteins such as prion protein (PrPC) and $\alpha$-synuclein. Maintaining adequate levels of both B12 and iron may therefore support neurotransmitter function and brain health, providing a nutritional avenue to indirectly reduce ASUD risk. 

Lidocaine, a sodium channel blocker, selectively attenuates cue-induced cocaine-seeking behavior in preclinical models by modulating amygdala circuits, though clinical studies have yet to show significant reductions in craving \cite{becker20,kantak02}. Hydroxyzine, particularly when combined with the 5-HT3 antagonist palonosetron, has been shown to alleviate opioid withdrawal severity, potentially lowering relapse risk \cite{erlendson17}. Lamotrigine, a glutamate-modulating antiepileptic, reduces cue-induced alcohol seeking and cocaine use in both preclinical and clinical settings through modulation of glutamatergic, dopaminergic, and serotonergic neurotransmission \cite{erlendson17,brown12}. Escitalopram and duloxetine, selective serotonin and serotonin/norepinephrine reuptake inhibitors, respectively, have demonstrated reductions in alcohol consumption and craving, with duloxetine also mitigating anxiety-like behaviors that may trigger relapse \cite{skelly14,mohammadi19}. Vitamin D (ergocalciferol) supplementation helps in modulating neurotransmitter systems, improving psychological symptoms, and reducing drug-seeking behaviors in at-risk populations \cite{ghaderi20,jalilian24}. Finally, ceftriaxone restores glutamate homeostasis via upregulation of GLT-1, attenuating cocaine- and alcohol-related reinstatement and partially reversing alcohol-induced gut dysbiosis, thereby further reducing vulnerability to ASUD \cite{duclot24,stennett17,rao14}. 
\begin{table}[!t]
\centering
\scriptsize 
\resizebox{155mm}{!}{
\renewcommand{\arraystretch}{1.2}
\begin{tabularx}{\textwidth}{
  >{\raggedright\arraybackslash}p{6.6cm}
  >{\raggedleft\arraybackslash}p{3.4cm}
  >{\raggedleft\arraybackslash}p{1.5cm}
  >{\centering\arraybackslash}p{2.4cm}
}
\toprule
\textbf{Indicators of Low Risk} & \textbf{Risk Marker Type} & \textbf{RC} & \textbf{N (Case/Ctrl)} \\
\midrule
Ferritin [Mass/volume] in Serum/Plasma & Abnormal Lab test result & 0.4231 & 48/108 \\
Vitamin B12 [Mass/volume] in Serum/Plasma & Abnormal Lab test result & 0.5072 & 70/86 \\
\midrule
Lidocaine & Medication use & 0.0121 & 210/1248\\
Hydroxyzine & Medication use & 0.0838 & 152/529\\
Lamotrigine & Medication use & 0.1219 & 73/240\\
Escitalopram & Medication use & 0.2128 & 78/407\\
Duloxetine & Medication use & 0.5460 & 111/489\\
Ergocalciferol & Medication use & 0.6546 & 40/222\\
Ceftriaxone & Medication use & 0.7809 & 51/168\\
\midrule
Disorder of Continuity of Bone & Comorbidities & 0.0010 & 35/ 87 \\
Other Disorders of Ear & Comorbidities & 0.0064 & 104/393 \\
COVID-19 & Comorbidities & 0.0086 & 120/598 \\
Other Speech Disturbances & Comorbidities & 0.0287 & 32/93 \\
\bottomrule
\end{tabularx}
}
\caption{\textbf{Key Indicators Associated with Decreased Risk of ASUD in Patients with PTSD.}}
\label{table6}
\end{table}

Certain comorbid conditions may confer a protective effect, through mechanisms related to increased healthcare engagement and structured medical supervision. Our analysis identified associations with reduced ASUD risk for disorders of bone continuity, other unspecified ear disorders, COVID-19, and unspecified speech disturbances. These conditions typically necessitate regular medical monitoring, specialist care, or ongoing management, providing opportunities to reinforce adaptive coping strategies, identify early signs of substance misuse, and limit exposure to high-risk behaviors. Consequently, such comorbidities may indirectly mitigate ASUD risk in PTSD patients by promoting treatment adherence and engagement in structured care environments.

\section*{Discussion}
In our study, we introduced BiPETE, a transformer-encoder classifier designed to predict single disease using structured, longitudinal EHR data. The central innovation of BiPETE lies in its bi-positional encoding strategy, which combines RoPE for capturing relative time gaps and SPE for encoding absolute visit order. This dual encoding captures fine-grained temporal dependencies in patient EHR sequence, addressing one of the key challenges in EHR modeling. The AUPRC scores of the variant models configured with only SPE or only RoPE indicate that RoPE plays a key role in modeling interdependencies among EHR tokens within and between visits, thereby enhancing the attention mechanism’s ability to focus on tokens relevant for distinguishing cases from controls (Table 1). However, the training and validation metrics of the RoPE only model (Figure 2) indicate instability, likely due to the variability of the days-ago indices across instances, which hinders learning of consistent temporal patterns. In contrast, the SPE-only model shows stable training but lower performance. By integrating SPE, which encodes absolute visit order, with RoPE, which models relative token relationships, BiPETE outperforms the single positional embedding variant models whilst retaining stability in model performance. Moreover, incorporating the most recent and clinically relevant patient visit data provided sufficient contextual information for BiPETE to identify label-specific patterns. These results demonstrate that optimizing positional encoding and incorporating task-relevant data substantially enhance the classification performance of transformer-based models.
 
BiPETE, which does not rely on pretraining, exhibited strong performance in distinguishing cases and controls–achieving over 90\% on both AUROC and AUPRC–across the depression and PTSD cohorts with sizes of 65,643 and 9,310, respectively. This suggests that the model identified distinct token patterns for the classes without learning the semantic meaning of the tokens. It is unlikely that the model had learned meaningful token embedding during training for the classification task alone. Model learning was facilitated by reducing diagnosis vocabulary size by over 90\% in both depression and PTSD cohorts. Without this step, ICD10 codes would dominate the vocabulary, leading to issues such as attention skew toward diagnosis, reduced representation of other vocabularies, and overfitting to ICD10 codes, all of which could impair model learning and generalizability. This process removed redundancy in diagnosis codes, enabling the model to learn meaning token patterns while integrating multiple vocabularies. We believe that BiPETE’s disease prediction performance could be further improved through pretraining on large-scale EHR corpora using tasks such as masked language modeling or disease prediction tasks to learn token context and semantics, as demonstrated in previous studies \cite{rasmy21,li20a,lindhagen15}; we leave this for the future work.

Our IG analysis provided key information about the biological and clinical markers associated with increased or reduced risk of ASUD among patients with depression and PTSD. In PTSD, elevated inflammatory and metabolic markers such as mean platelet volume and hypoalbuminemia suggested systemic stress and inflammation as key pathways linking trauma-related dysregulation to substance misuse. Concurrently, medications such as hydrocodone and oxybutynin highlight the complex interaction between pain management, neuropsychiatric effects, and addiction risk. Conversely, protective features including adequate vitamin B12 and ferritin levels, and use of agents like lamotrigine and ceftriaxone, point toward neurobiological stabilization and glutamatergic modulation as potential resilience mechanisms. Similar patterns emerged in the depression cohort, where immune and coagulation markers (e.g., altered NLR, PT/INR) reflected shared inflammatory and hepatic pathways underlying comorbid depression and SUD. Notably, $\beta$-lactam antibiotics and psychostimulant therapies appeared protective, consistent with emerging evidence for neuroimmune and dopaminergic modulation in addiction recovery. Collectively, these findings demonstrate how BiPETE can bridge data-driven prediction with mechanistic understanding, revealing clinically meaningful targets for early identification and personalized intervention in comorbid psychiatric and substance use disorders.

There are several limitations in our current work that should be addressed in future evaluations of BiPETE. First, we tested BiPETE only on the ASUD risk prediction task, so its performance on other disease prediction tasks remains uncertain. Second, BiPETE has not yet been evaluated on datasets beyond the AoU data, and its generalizability remains to be fully validated. Third, without pretraining to learn token contextual information, BiPETE is currently limited to single-disease prediction and requires retraining for each new disease prediction task. Fourth, we did not include any demographic information in our input. Incorporating such information could improve performance, especially given the diverse racial composition of our cohort (Supplementary Table 1). In addition, there are methodological caveats in interpreting attributions from transformer models. The context-dependent nature of the attention mechanism can lead to varying attributions for the same code across patients or visits. Furthermore, confounding by indication may cause treatment codes associated with more severe disease to appear linked to ASUD outcomes, even when they reflect disease progression rather than risk.

In conclusion, we proposed a single-disease classifier, BiPETE, and evaluated it on ASUD risk prediction. BiPETE’s positional encoding configuration could be applied to other clinical relevant tasks by improving temporal modeling of EHR data. This configuration can be readily adapted to existing transformer-based models to potentially boost their performance. BiPETE is a high-performing classifier, making it particularly valuable in settings where pretrained models or large-scale pretraining data are unavailable.
\clearpage

\section*{Acknowledgments}
We gratefully acknowledge All of Us participants for their contributions, without whom this research would not have been possible. We also thank the National Institutes of Health’s All of Us Research Program for making available the participant data and cohorts examined in this study.

\clearpage

\section*{Supplementary Information}

\begin{table}[!ht]
\small
\centering
\resizebox{155mm}{!}{
\renewcommand{\arraystretch}{1.2}
\noindent\begin{tabular}{
   >{\centering}p{40mm}
   >{\centering}p{5.75cm}
   >{\centering\arraybackslash}p{5.75cm}
}
\toprule
\multicolumn{1}{c}{\bfseries Cohort Stats} 
& \multicolumn{1}{c}{\bfseries Depressive Disorder} 
& \multicolumn{1}{c}{\bfseries PTSD} \\
\midrule
COHORT SIZE &  84,163 & 15,334 \\
\midrule
CASES\%:CTRL\% & 21.3\% : 78.7\% & 43.9\% : 56.1\% \\
\midrule
GENDER RATIO\\M\%:F\% & \raisebox{-0.8\height}{26.5\% : 73.5\%} & \raisebox{-0.8\height}{35.4\% : 64.6\%}\\
\midrule
\raisebox{-0.9\height}{\shortstack{AGE\\DISTRIBUTION}}& 
    \begin{tabular}[t]{
         >{\centering}p{1.5cm}
         >{\centering}p{1.5cm}
         >{\centering\arraybackslash}p{1.5cm}
      }
      \multicolumn{1}{c}{18-44} 
      & \multicolumn{1}{c}{45-66} 
      & \multicolumn{1}{c}{$\geq$66}\\
      \midrule
      39.3\% & 43.6\% & 18.1\%\\
      \end{tabular} \\
    &
    \begin{tabular}[t]{
     >{\centering}p{1.5cm}
     >{\centering}p{1.5cm}
     >{\centering\arraybackslash}p{1.5cm}
    }
    \multicolumn{1}{c}{18-44} 
    & \multicolumn{1}{c}{45-66} 
    & \multicolumn{1}{c}{$\geq$66}\\
    \midrule
    32.2\% & 39.9\% & 27.9\%\\
    \end{tabular} \\
\midrule
WHITE & 54\% & 62\%\\
AFAM & 2\% & 14\%\\
AI/AN & 2\% & 1\%\\
ASIAN & 1\% & 2\%\\
MENA & $<$1\% & $<$1\% \\
NHPI & $<$1\% & $<$1\% \\
OTHER & 25\% & 21\%\\      
& & \\[-2.7ex]
\bottomrule
\end{tabular}
}
\caption*{\textbf{Supplementary Table 1. Cohort Characteristics Preceding Data Preprocessing.} Gender reflects sex assigned at birth. Age distribution is based on participants' age at the time of consent to data collection for the AoU program. The "Other" category in the race row includes individuals who did not answer, selected multiple populations, or indicated a race not listed. Abbreviations of Race: AFAM–African American; AI/AN–American Indian or Alaska Native; MENA–Middle Eastern or North African; NHPI–Native Hawaiian or Other Pacific Islander.}
\label{suptable1}
\end{table}
\clearpage

\begin{figure}[t!]
  \centering
  \includegraphics[width=1\textwidth]{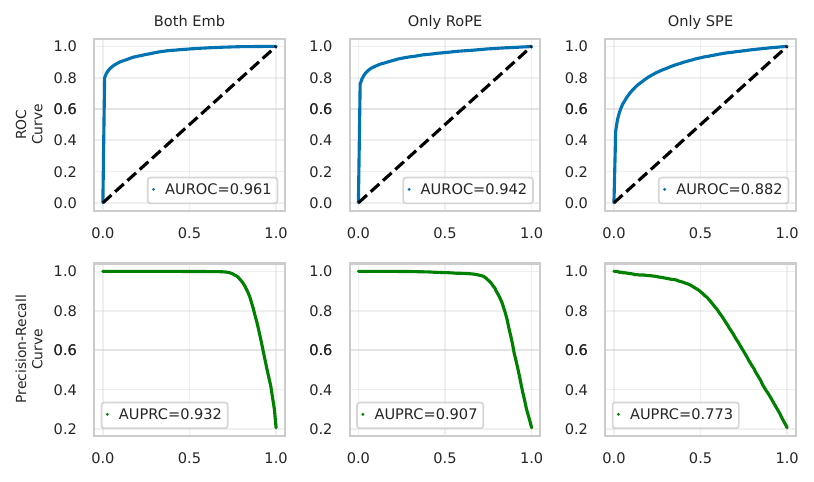}
  \caption*{\textbf{Supplementary Figure 1.} \textbf{Test ROC and PR Curve Comparison of Classifiers with Different Positional Encoding Configurations.} For receiver operating characteristic (ROC) curves, the x-axis shows the false positive rate, and the y-axis shows the true positive rate. For precision-recall curves, the x-axis shows recall, and the y-axis shows precision. Curves and metrics are computed using the mean values of the corresponding rates across the cross-validation folds.   }
  \label{supfig1}
\end{figure}
\clearpage

\begin{table}[t!]
\centering
\small
\resizebox{145mm}{!}{
\renewcommand{\arraystretch}{1.3}
\begin{tabular}{llcc >{\centering\arraybackslash}p{0.65cm} c >{\centering\arraybackslash}p{0.65cm} c >{\centering\arraybackslash}p{0.65cm}}\\
\toprule
&\textbf{Model}&\textbf{T} & \textbf{PPV} & \textbf{CV} & \textbf{Sensitivity} & \textbf{CV} & \textbf{Specificity} & \textbf{CV} \\
\hline
\multirow{9}{*}{\textbf{Dep}}
&\multirow{3}{*}{\textbf{BiPETE}}& 0.2 & $91.42 \; (\pm2.06)$ & \multirow{3}{*}{1.41} & $82.59 \; (\pm 1.41)$ & \multirow{3}{*}{1.19} & $97.97 \; (\pm 0.56)$ & \multirow{3}{*}{0.35}\\
& & 0.5 & $93.28 \; (\pm2.04)$ & & $81.36 \; (\pm1.71)$ & & $98.46 \; (\pm0.53)$ &\\
& & 0.8 & $94.61 \; (\pm1.93)$ & & $80.22 \; (\pm1.85)$ & & $98.80 \; (\pm0.48)$ &\\
\cline{2-9}
&\multirow{3}{*}{\textbf{BiGRU}}& 0.2 & $67.19 \; (\pm5.81)$ &\multirow{3}{*}{4.34} & $55.23 \; (\pm 5.56)$ &\multirow{3}{*}{6.25} & $92.69 \; (\pm 2.60)$ &\multirow{3}{*}{1.35}\\
& & 0.5 & $71.05 \; (\pm5.81)$ & & $51.23 \; (\pm5.42)$ & & $94.32 \; (\pm2.14)$ &\\
& & 0.8 & $74.74 \; (\pm5.59)$ & & $47.38 \; (\pm5.30)$ & & $95.64 \; (\pm1.73)$ &\\
\cline{2-9}
&\multirow{3}{*}{\textbf{LR}}& 0.2 & $45.42 \; (\pm0.22)$ &\multirow{3}{*}{25.70} & $74.86 \; (\pm 0.92)$ &\multirow{3}{*}{35.28} & $76.71 \; (\pm 0.24)$ &\multirow{3}{*}{10.76}\\
& & 0.5 & $72.61 \; (\pm0.31)$ & & $49.80 \; (\pm0.56)$ & & $95.14 \; (\pm0.09)$ &\\
& & 0.8 & $88.17 \; (\pm1.21)$ & & $30.33 \; (\pm0.68)$ & & $98.94 \; (\pm0.14)$ &\\
\cline{2-9}
\hline
\multirow{9}{*}{\textbf{PTSD}}
&\multirow{3}{*}{\textbf{BiPETE}}& 0.2 & $88.48 \; (\pm3.47)$ &\multirow{3}{*}{1.08} & $85.50 \; (\pm 2.31)$ &\multirow{3}{*}{0.52} & $96.28 \; (\pm 1.34)$ &\multirow{3}{*}{0.37}\\
& & 0.5 & $89.40 \; (\pm3.36)$ & & $85.02 \; (\pm2.28)$ & & $96.62 \; (\pm1.34)$ &\\
& & 0.8 & $90.83 \; (\pm3.24)$ & & $84.41 \; (\pm2.41)$ & & $97.15 \; (\pm1.14)$ &\\
\cline{2-9}
&\multirow{3}{*}{\textbf{BiGRU}}& 0.2 & $60.60 \; (\pm4.68)$ &\multirow{3}{*}{5.16} & $52.09 \; (\pm 5.75)$ &\multirow{3}{*}{9.80} & $88.52 \; (\pm 3.22)$ &\multirow{3}{*}{2.28}\\
& & 0.5 & $63.89 \; (\pm4.26)$ & & $45.99 \; (\pm6.09)$ & & $91.19 \; (\pm2.58)$ &\\
& & 0.8 & $68.69 \; (\pm4.73)$ & & $40.98 \; (\pm6.37)$ & & $93.61 \; (\pm2.22)$ &\\
\cline{2-9}
&\multirow{3}{*}{\textbf{LR}}& 0.2 & $49.45 \; (\pm1.35)$ &\multirow{3}{*}{20.37} & $78.40 \; (\pm 1.23)$ &\multirow{3}{*}{25.12} & $73.72 \; (\pm 1.57)$ &\multirow{3}{*}{11.34}\\
& & 0.5 & $67.65 \; (\pm1.46)$ & & $58.45 \; (\pm1.14)$ & & $90.83 \; (\pm0.72)$ &\\
& & 0.8 & $82.61 \; (\pm1.77)$ & & $41.81 \; (\pm2.83)$ & & $97.12 \; (\pm0.34)$ &\\
\bottomrule
\end{tabular}
}
\caption*{\textbf{Supplementary Table 2. Performance Metrics for ASUD Risk Prediction using BiPETE and Baseline Models Across Decision Thresholds in Depression and PTSD Cohorts.} PPV (Precision), Sensitivity (Recall) and Specificity (True Negative Rate) are reported for comparisons with baseline models. Metrics are calculated at three decision thresholds: 0.2, 0.5 and 0.8. ± indicates standard deviation across five-fold cross-validation. Coefficient of Variation (CV), expressed as a percentage, is reported for each metric to show its variability across thresholds.}
\label{suptable2}
\end{table}
\clearpage

\begin{figure}[t!]
  \centering
  \includegraphics[width=1\textwidth]{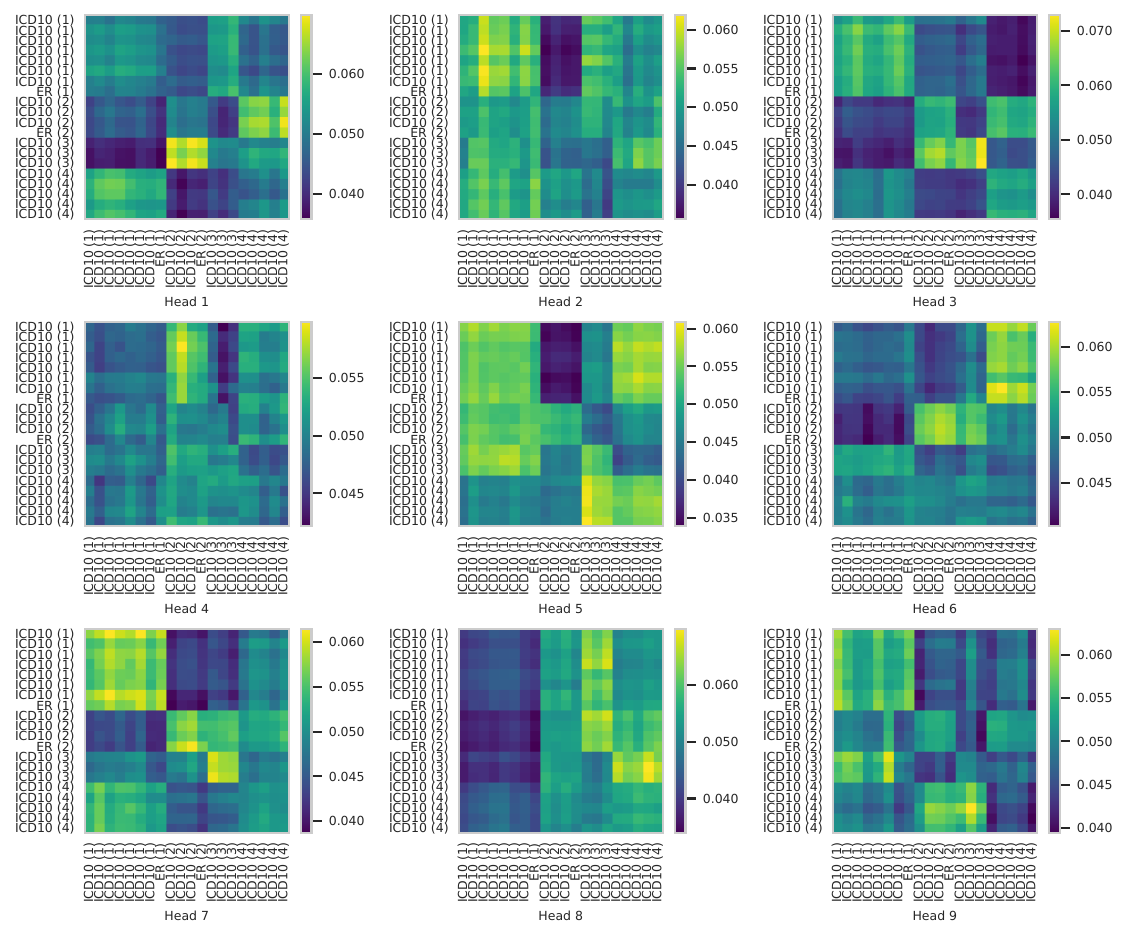}
  \caption*{\textbf{Supplementary Figure 2. Attention Heads of Final Layer in BiPETE.} The attention is calculated using a representative EHR code sequence of length 20. EHR codes in axes labels are replaced with the vocabulary type to preserve patient anonymity. Attention maps show tokens tend to receive higher or lower scores in groups of tokens from the same visit. Different heads focus on distinct visit-level token groups, seeking different patterns in the sequence.}
  \label{supfig2}
\end{figure}

\begin{table}[H]
\centering
\label{tab:aou-asud}
{\fontsize{7.5pt}{8pt}\selectfont
\begin{tabularx}{\linewidth}{|Y|L{1.6cm}|Y|L{1.6cm}|}
\hline
\textbf{Depressive Disorder Concept Set} & \textbf{Concept Id} & \textbf{PTSD Concept Set} & \textbf{Concept Id} \\
\hline
Depressive disorder & 440383 & Acute post-trauma stress state & 4100536\\
 & & Chronic post-traumatic stress disorder & 443414\\
 & & Complex posttraumatic stress disorder & 40484109\\
 & & Posttraumatic stress disorder & 436676\\
\hline
\end{tabularx}}
\caption*{\textbf{Supplementary Table 3.} All of Us Concept Sets Used to Define MHD Cohorts.}
\label{suptable3}
\end{table}

\begin{table}[ht]
\centering
{\fontsize{7.5pt}{8pt}\selectfont
\begin{tabularx}{\linewidth}{|Y|L{1.6cm}|Y|L{1.6cm}|}
\hline
\textbf{ASUD Concept Set} & \textbf{Concept Id} & \textbf{ASUD Concept Set (Continued)} & \textbf{Concept Id}\\
\hline
Abuse of antidepressant drug & 440992 & Drug dependence in remission & 43530680\\ 
Accidental poisoning by hallucinogens & 438946 & Drug withdrawal & 441260\\ 
Accidental poisoning by heroin & 440307 & Drug-induced amnestic syndrome & 373172\\ 
Acute alcoholic intoxication in alcoholism & 433735 & Drug-induced delirium & 373449\\ 
Acute alcoholic intoxication in remission, in alcoholism & 432609 & Drug-induced delusional disorder & 443559\\ 
Acute alcoholic liver disease & 201343 & Drug-induced dementia & 376095\\ 
Alcohol abuse & 433753 & Drug-induced paranoia or hallucinatory states & 4101142\\ 
Alcohol dependence & 435243 & Drug-induced psychosis & 434900\\ 
Alcohol withdrawal syndrome & 375519 & Drug-induced sleep disorder & 435792\\ 
Alcohol-induced anxiety disorder & 4146660 & Episodic acute alcoholic intoxication in alcoholism & 441261\\ 
Alcohol-induced mood disorder & 4205002 & Gastric hemorrhage due to alcoholic gastritis & 45757783\\ 
Alcohol-induced psychotic disorder with delusions & 442582 & Hallucinogen abuse & 437245\\ 
Alcohol-induced sleep disorder & 375794 & Hallucinogen dependence & 433180\\ 
Alcoholic fatty liver & 193256 & Hallucinogen dependence in remission & 434921\\ 
Alcoholic gastritis & 195300 & Hallucinosis caused by drug & 440987\\ 
Amphetamine abuse & 432878 & Hypnotic or anxiolytic abuse & 439554\\ 
Amphetamine and amphetamine derivative drug dependence & 3654785 & Inhalant abuse & 4290538\\ 
Amphetamine dependence & 437533 & Inhalant dependence & 4176120\\ 
Amphetamine or psychostimulant dependence in remission & 432884 & Inhalant-induced mood disorder & 4232492\\ 
Amphetamine or psychostimulant dependence, continuous & 434916 & Inhalant-induced organic mental disorder & 4264889\\ 
Amphetamine or psychostimulant dependence, episodic & 441262 & Mood disorder caused by drug & 436079\\ 
Cannabis abuse & 434327 & Nicotine dependence & 4209423\\ 
Cannabis dependence & 440387 & Nicotine dependence in remission & 3654548\\ 
Cannabis dependence in remission & 440996 & Nondependent antidepressant type drug abuse in remission & 439313\\ 
Cannabis intoxication delirium & 4220197 & Nondependent cocaine abuse in remission & 436098\\ 
Cannabis-induced anxiety disorder & 4221077 & Nondependent cocaine abuse, continuous & 439796\\ 
Cannabis-induced psychotic disorder with hallucinations & 4097389 & Nondependent hallucinogen abuse & 4150794\\ 
Cocaine abuse & 432303 & Nondependent hallucinogen abuse in remission & 441272\\ 
Cocaine dependence & 436389 & Nondependent mixed drug abuse & 439312\\ 
Cocaine dependence in remission & 432302 & Nondependent mixed drug abuse in remission & 4103426\\ 
Cocaine-induced anxiety disorder & 4198826 & Nondependent opioid abuse & 4099935\\ 
Cocaine-induced mood disorder & 4012869 & Nondependent opioid abuse in remission & 436088\\ 
Cocaine-induced psychotic disorder with hallucinations & 4272033 & Opioid abuse & 438130\\ 
Combined drug dependence, excluding opioid, in remission & 433458 & Opioid dependence & 438120\\ 
Combined drug dependence, excluding opioids & 436370 & Opioid dependence in remission & 432301\\ 
Combined opioid with other drug dependence & 4099809 & Poisoning by heroin & 433919\\ 
Combined opioid with other drug dependence in remission & 4100520 & Poisoning by methadone & 440919\\ 
Combined opioid with other drug dependence, continuous & 4102817 & Poisoning by opium alkaloid & 439223\\ 
Combined opioid with other drug dependence, episodic & 4103413 & Psychoactive substance abuse & 4239381\\ 
Continuous acute alcoholic intoxication in alcoholism & 437257 & Psychoactive substance dependence & 4080762\\ 
Delirium due to sedative withdrawal & 4262566 & Psychoactive substance use disorder & 4004672\\ 
Dementia associated with alcoholism & 378726 & Psychoactive substance-induced organic hallucinosis & 4155336\\ 
Dilated cardiomyopathy secondary to alcohol & 318773 & Psychotic disorder caused by cocaine & 37110437\\ 
Disorder caused by alcohol & 36714559 & Stimulant abuse & 40479573\\ 
Drug dependence & 440069 & Therapeutic drug dependence & 4319165\\ 
Drug dependence during pregnancy - baby delivered & 442915 & & \\ 
Drug dependence in mother complicating pregnancy, childbirth AND/OR puerperium & 440787 & & \\
\hline
\end{tabularx}}
\caption*{\textbf{Supplementary Table 4.} All of Us Concept Set Used to Define ASUD.}
\label{suptable4}
\end{table}


\clearpage

\begin{thebibliography}{00}

\bibitem{rajpurkar22}
  Rajpurkar, P., Chen, E., Banerjee, O. \& Topol, E. J.,
  \textit{AI in health and medicine},
  Nature medicine,
  28, 31-38,
  (2022).

\bibitem{hama25}
  Hama, T. et al.,
  \textit{Enhancing patient outcome prediction through deep learning with sequential diagnosis codes from structured electronic health record data: Systematic review},
  Journal of Medical Internet Research 27, e57358, (2025).

\bibitem{rasmy22}
  Rasmy, L. et al.,
  \textit{Recurrent neural network models (CovRNN) for predicting outcomes of patients with COVID-19 on admission to hospital: model development and validation using electronic health record data},
  The Lancet Digital Health 4, e415--e425, (2022).

\bibitem{atasoy19}
  Atasoy, H., Greenwood, B. N. \& McCullough, J. S.,
  \textit{The digitization of patient care: a review of the effects of electronic health records on health care quality and utilization},
  Annual Review of Public Health 40, 487--500, (2019).

\bibitem{yangx22}
  Yang, X. et al.,
  \textit{A large language model for electronic health records},
  NPJ Digital Medicine 5, 194, (2022).

\bibitem{rasmy21}
  Rasmy, L., Xiang, Y., Xie, Z., Tao, C. \& Zhi, D.,
  \textit{Med-BERT: pretrained contextualized embeddings on large-scale structured electronic health records for disease prediction},
  NPJ Digital Medicine 4, 86, (2021).

\bibitem{devlin19}
  Devlin, J., Chang, M.-W., Lee, K. \& Toutanova, K.,
  \textit{BERT: Pre-training of deep bidirectional transformers for language understanding},
  in Proceedings of the 2019 Conference of the North American Chapter of the Association for Computational Linguistics (NAACL-HLT), 4171--4186, (2019).

\bibitem{xie22}
  Xie, F. et al.,
  \textit{Deep learning for temporal data representation in electronic health records: A systematic review of challenges and methodologies},
  Journal of Biomedical Informatics 126, 103980, (2022).

\bibitem{holmes21}
  Holmes, J. H. et al.,
  \textit{Why is the electronic health record so challenging for research and clinical care?},
  Methods of Information in Medicine 60, 032--048, (2021).

\bibitem{li20a}
  Li, Y. et al.,
  \textit{BEHRT: transformer for electronic health records},
  Scientific Reports 10, 1--12, (2020).

\bibitem{si21}
  Si, Y. et al.,
  \textit{Deep representation learning of patient data from Electronic Health Records (EHR): A systematic review},
  Journal of Biomedical Informatics 115, 103671, (2021).

\bibitem{lindhagen15}
  Lindhagen, L., Van Hemelrijck, M., Robinson, D., Stattin, P. \& Garmo, H.,
  \textit{How to model temporal changes in comorbidity for cancer patients using prospective cohort data},
  BMC Medical Informatics and Decision Making 15, 96, (2015).

\bibitem{valderas09}
  Valderas, J. M., Starfield, B., Sibbald, B., Salisbury, C. \& Roland, M.,
  \textit{Defining comorbidity: implications for understanding health and health services},
  The Annals of Family Medicine 7, 357--363, (2009).

\bibitem{yangz23}
  Yang, Z., Mitra, A., Liu, W., Berlowitz, D. \& Yu, H.,
  \textit{TransformEHR: transformer-based encoder-decoder generative model to enhance prediction of disease outcomes using electronic health records},
  Nature Communications 14, 7857, (2023).

\bibitem{su24}
  Su, J. et al.,
  \textit{Roformer: Enhanced transformer with rotary position embedding},
  Neurocomputing 568, 127063, (2024).

\bibitem{rabbani22}
  Rabbani, N., Kim, G. Y., Suarez, C. J. \& Chen, J. H.,
  \textit{Applications of machine learning in routine laboratory medicine: Current state and future directions},
  Clinical Biochemistry 103, 1--7, (2022).

\bibitem{bailly22}
  Bailly, A. et al.,
  \textit{Effects of dataset size and interactions on the prediction performance of logistic regression and deep learning models},
  Computer Methods and Programs in Biomedicine 213, 106504, (2022).

\bibitem{ramirez22}
  Ramirez, A. H. et al.,
  \textit{The All of Us Research Program: data quality, utility, and diversity},
  Patterns 3, (2022).

\bibitem{sadeghi24}
  Sadeghi, Z. et al.,
  \textit{A review of Explainable Artificial Intelligence in healthcare},
  Computers and Electrical Engineering 118, 109370, (2024).

\bibitem{sundararajan17}
  Sundararajan, M., Taly, A. \& Yan, Q.,
  \textit{Axiomatic attribution for deep networks (Integrated Gradients)} in International Conference on Machine Learning (ICML), 3319--3328 (PMLR), (2017).
  
\bibitem{abgrall24}
  Abgrall, G., Holder, A. L., Chelly Dagdia, Z., Zeitouni, K. \& Monnet, X.,
  \textit{Should AI models be explainable to clinicians?},
  Critical Care 28, 301, (2024).
  
\bibitem{rodriguez09}
  Rodriguez JD, Perez A, \& Lozano JA,
  \textit{Sensitivity analysis of k-fold cross validation in
prediction error estimation}, IEEE transactions on pattern analysis and machine intelligence 32, 569-575, (2009).
  
\bibitem{hoffmann22}
  Hoffmann, J.et al., \textit{Training Compute-Optimal Large Language Models} in NeurIPS 22: Proceedings of the 36th International Conference on Neural Information Processing Systems, 30016–30030, (2022). 
https://doi.org/10.48550/arXiv.2203.15556
  

\bibitem{martin23}
  Mart\'in-Gonz\'alez, C. et al.,
  \textit{Mean platelet volume and mortality in patients with alcohol use disorder},
  Digestive and Liver Disease 55, 1236--1241, (2023).

\bibitem{annoni90}
  Annoni, G., Weiner, F. R., Colombo, M., Czaja, M. J. \& Zern, M. A.,
  \textit{Albumin and collagen gene regulation in alcohol- and virus-induced human liver disease},
  Gastroenterology 98, 197--202, (1990).

\bibitem{darwish17}
  Darwish, M. et al.,
  \textit{Abuse potential with oral route of administration of a hydrocodone extended-release tablet formulated with abuse-deterrence technology in nondependent, recreational opioid users},
  Pain Medicine 18, 61--77, (2017).

\bibitem{miller04}
  Miller, N. S. \& Greenfeld, A.,
  \textit{Patient characteristics and risks factors for development of dependence on hydrocodone and oxycodone},
  American Journal of Therapeutics 11, 26--32, (2004).

\bibitem{schwartz06}
  Schwartz, A. C. et al.,
  \textit{Pain medication use among patients with posttraumatic stress disorder},
  Psychosomatics 47, 136--142, (2006).

\bibitem{gulsun06}
  Gulsun, M., Pinar, M. \& Sabanci, U.,
  \textit{Psychotic disorder induced by oxybutynin: Presentation of two cases},
  Clinical Drug Investigation 26, 603--606, (2006).

\bibitem{welk22}
  Welk, B., Etaby, K., McArthur, E. \& Chou, Q.,
  \textit{The risk of delirium and falls or fractures with the use of overactive bladder anticholinergic medications},
  Neurourology and Urodynamics 41, 348--356, (2022).

\bibitem{iqbal22}
  Iqbal, K. M., Luke, P. K. \& Ingram, M. T.,
  \textit{Psychosis resulting from trimethoprim-sulfamethoxazole treatment for preseptal cellulitis},
  Taiwan Journal of Ophthalmology 12, 223--226, (2022).

\bibitem{stuhec14}
  Stuhec, M.,
  \textit{Trimethoprim-sulfamethoxazole-related hallucinations},
  General Hospital Psychiatry 36, 230.e237--230.e238, (2014).

\bibitem{scholz24}
  Scholz, S. M., Thalmann, N. F., M\"uller, D., Trippolini, M. A. \& Wertli, M. M.,
  \textit{Factors influencing pain medication and opioid use in patients with musculoskeletal injuries: a retrospective insurance claims database study},
  Scientific Reports 14, 1978, (2024).

\bibitem{pastore18}
  Pastore, G. P., Goulart, D. R., Pastore, P. R., Prati, A. J. \& de Moraes, M.,
  \textit{Self-medication among myofascial pain patients: a preliminary study},
  The Open Dentistry Journal 12, 347, (2018).

\bibitem{berlt25}
  Berlt, M., de Souza, K. B., Zhang, L., Bock, P. M. \& Hort, M. A.,
  \textit{Prevalence of self-medication for dental issues in the general population: a systematic review and meta-analysis},
  Discover Public Health 22, 1--32, (2025).

\bibitem{becker20}
  Becker, J. E. et al.,
  \textit{The efficacy of lidocaine in disrupting cocaine cue-induced memory reconsolidation},
  Drug and Alcohol Dependence 212, 108062, (2020).

\bibitem{kantak02}
  Kantak, K. M., Black, Y., Valencia, E., Green-Jordan, K. \& Eichenbaum, H. B.,
  \textit{Dissociable effects of lidocaine inactivation of the rostral and caudal basolateral amygdala on the maintenance and reinstatement of cocaine-seeking behavior in rats},
  Journal of Neuroscience 22, 1126--1136, (2002).

\bibitem{erlendson17}
  Erlendson, M. J. et al.,
  \textit{Palonosetron and hydroxyzine pre-treatment reduces the objective signs of experimentally-induced acute opioid withdrawal in humans: a double-blinded, randomized, placebo-controlled crossover study},
  The American Journal of Drug and Alcohol Abuse 43, 78--86, (2017).

\bibitem{vengeliene07}
  Vengeliene, V., Heidbreder, C. A. \& Spanagel, R.,
  \textit{The effects of lamotrigine on alcohol seeking and relapse},
  Neuropharmacology 53, 951--957, (2007).

\bibitem{brown12}
  Brown, E. S., Sunderajan, P., Hu, L. T., Sowell, S. M. \& Carmody, T. J.,
  \textit{A randomized, double-blind, placebo-controlled, trial of lamotrigine therapy in bipolar disorder, depressed or mixed phase and cocaine dependence},
  Neuropsychopharmacology 37, 2347--2354, (2012).

\bibitem{skelly14}
  Skelly, M. J. \& Weiner, J. L.,
  \textit{Chronic treatment with prazosin or duloxetine lessens concurrent anxiety-like behavior and alcohol intake: evidence of disrupted noradrenergic signaling in anxiety-related alcohol use},
  Brain and Behavior 4, 468--483, (2014).

\bibitem{mohammadi19}
  Mohammadi, N. et al.,
  \textit{Preventive effects of duloxetine against methamphetamine induced neurodegeneration and motor activity disorder in rat: Possible role of CREB/BDNF signaling pathway},
  International Journal of Preventive Medicine 10, 195, (2019).

\bibitem{ghaderi20}
  Ghaderi, A. et al.,
  \textit{Exploring the effects of vitamin D supplementation on cognitive functions and mental health status in subjects under methadone maintenance treatment},
  Journal of Addiction Medicine 14, 18--25, (2020).

\bibitem{jalilian24}
  Jalilian-Khave, L. et al.,
  \textit{Potential roles for vitamin D in preventing and treating impulse control disorders, behavioral addictions, and substance use disorders: A scoping review},
  Addiction Neuroscience, 100190, (2024).

\bibitem{duclot24}
  Duclot, F., Wu, L., Wilkinson, C. S., Kabbaj, M. \& Knackstedt, L. A.,
  \textit{Ceftriaxone alters the gut microbiome composition and reduces alcohol intake in male and female Sprague--Dawley rats},
  Alcohol 120, 169--178, (2024).

\bibitem{stennett17}
  Stennett, B. A., Frankowski, J. C., Peris, J. \& Knackstedt, L. A.,
  \textit{Ceftriaxone reduces alcohol intake in outbred rats while upregulating xCT in the nucleus accumbens core},
  Pharmacology Biochemistry and Behavior 159, 18--23, (2017).

\bibitem{rao14}
  Rao, P. \& Sari, Y.,
  \textit{Effectiveness of ceftriaxone treatment in preventing relapse-like drinking behavior following long-term ethanol dependence in P rats},
  Journal of Addiction Research \& Therapy 5, 1000183, (2014).

\bibitem{karatoprak21}
  Karatoprak, S., Uzun, N., Ak{\i}nc{\i}, M. A. \& D\"onmez, Y. E.,
  \textit{Neutrophil-lymphocyte and platelet-lymphocyte ratios among adolescents with substance use disorder: A preliminary study},
  Clinical Psychopharmacology and Neuroscience 19, 669, (2021).

\bibitem{gasparyan19}
  Gasparyan, A. Y., Ayvazyan, L., Mukanova, U., Yessirkepov, M. \& Kitas, G. D.,
  \textit{The platelet-to-lymphocyte ratio as an inflammatory marker in rheumatic diseases},
  Annals of Laboratory Medicine 39, 345, (2019).

\bibitem{sharma19}
  Sharma, T., Kumar, M., Rizkallah, A., Cappelluti, E. \& Padmanabhan, P.,
  \textit{Cocaine-induced thrombosis: review of predisposing factors, potential mechanisms, and clinical consequences with a striking case report},
  Cureus 11, (2019).

\bibitem{alonzo19}
  Alonzo, M. M., Lewis, T. V. \& Miller, J. L.,
  \textit{Disulfiram-like reaction with metronidazole: an unsuspected culprit},
  The Journal of Pediatric Pharmacology and Therapeutics 24, 445--449, (2019).

\bibitem{martinez11}
  Martinez-Diaz, G. J. \& Hsia, R.,
  \textit{Altered mental status from acyclovir},
  The Journal of Emergency Medicine 41, 55--58, (2011).

\bibitem{ng19}
  Ng, M.-H. et al.,
  \textit{Macrocytosis among patients with heroin use disorder},
  Neuropsychiatric Disease and Treatment, 2293--2298, (2019).

\bibitem{li24}
  Li, D. et al.,
  \textit{The relationship between mean corpuscular hemoglobin concentration and mortality in hypertensive individuals: A population-based cohort study},
  PLOS ONE 19, e0301903, (2024).

\bibitem{bohler23}
  Bohler, R. M. et al.,
  \textit{The policy landscape for naloxone distribution in four states highly impacted by fatal opioid overdoses},
  Drug and Alcohol Dependence Reports 6, 100126, (2023).

\bibitem{bazazi10}
  Bazazi, A. R., Zaller, N. D., Fu, J. J. \& Rich, J. D.,
  \textit{Preventing opiate overdose deaths: examining objections to take-home naloxone},
  Journal of Health Care for the Poor and Underserved 21, 1108--1113, (2010).

\bibitem{fischer25}
  Fischer, L. S. et al.,
  \textit{Effectiveness of naloxone distribution in community settings to reduce opioid overdose deaths among people who use drugs: a systematic review and meta-analysis},
  BMC Public Health 25, 1135, (2025).

\bibitem{petrovitch24}
  Petrovitch, D. et al.,
  \textit{State program enables the identification of factors associated with naloxone awareness, self-efficacy, and use for overdose reversal: A cross-sectional, observational study in an urban emergency department population},
  Journal of Substance Use and Addiction Treatment 167, 209506, (2024).

\bibitem{alasmari16}
  Alasmari, F., Rao, P. \& Sari, Y.,
  \textit{Effects of cefazolin and cefoperazone on glutamate transporter 1 isoforms and cystine/glutamate exchanger as well as alcohol drinking behavior in male alcohol-preferring rats},
  Brain Research 1634, 150--157, (2016).

\bibitem{rao15}
  Rao, P. et al.,
  \textit{Effects of ampicillin, cefazolin and cefoperazone treatments on GLT-1 expressions in the mesocorticolimbic system and ethanol intake in alcohol-preferring rats},
  Neuroscience 295, 164--174, (2015).

\bibitem{weiland15}
  Weiland, A., Garcia, S. \& Knackstedt, L. A.,
  \textit{Ceftriaxone and cefazolin attenuate the cue-primed reinstatement of alcohol-seeking},
  Frontiers in Pharmacology 6, 44, (2015).

\bibitem{mergenhagen20}
  Mergenhagen, K. A., Wattengel, B. A., Skelly, M. K., Clark, C. M. \& Russo, T. A.,
  \textit{Fact versus fiction: a review of the evidence behind alcohol and antibiotic interactions},
  Antimicrobial Agents and Chemotherapy 64, 10.1128/aac.02167-02119, (2020).

\bibitem{hakami16}
  Hakami, A. Y., Hammad, A. M. \& Sari, Y.,
  \textit{Effects of amoxicillin and augmentin on cystine-glutamate exchanger and glutamate transporter 1 isoforms as well as ethanol intake in alcohol-preferring rats},
  Frontiers in Neuroscience 10, 171, (2016).

\bibitem{callans24}
  Callans, L. S. et al.,
  \textit{Clavulanic acid decreases cocaine cue reactivity in addiction-related brain areas, a randomized fMRI pilot study},
  Psychopharmacology Bulletin 54, 8, (2024).

\bibitem{philogene22}
  Philogene-Khalid, H. L. et al.,
  \textit{The GLT-1 enhancer clavulanic acid suppresses cocaine place preference behavior and reduces GCPII activity and protein levels in the rat nucleus accumbens},
  Drug and Alcohol Dependence 232, 109306, (2022).

\bibitem{maser24}
  Maser, J. et al.,
  \textit{Clavulanic Acid-Mediated Increases in Anterior Cingulate Glutamate Levels are Associated With Decreased Cocaine Craving and Brain Network Functional Connectivity Changes},
  Current Therapeutic Research 101, 100751, (2024).

\bibitem{schroeder14}
  Schroeder, J. A. et al.,
  \textit{Clavulanic acid reduces rewarding, hyperthermic and locomotor-sensitizing effects of morphine in rats: a new indication for an old drug?},
  Drug and Alcohol Dependence 142, 41--45, (2014).

\bibitem{mather20}
  Mather, J. F., Seip, R. L. \& McKay, R. G.,
  \textit{Impact of famotidine use on clinical outcomes of hospitalized patients with COVID-19},
  Official journal of the American College of Gastroenterology | ACG 115, 1617--1623, (2020).

\bibitem{aouizerate06}
  Aouizerate, B. et al.,
  \textit{Glucocorticoid negative feedback in methadone-maintained former heroin addicts with ongoing cocaine dependence: dose--response to dexamethasone suppression},
  Addiction Biology 11, (2006).

\bibitem{capasso97}
  Capasso, A., Pinto, A., Sorrentino, L. \& Cirino, G.,
  \textit{Dexamethasone inhibition of acute opioid physical dependence in vitro is reverted by anti-lipocortin-1 and mimicked by anti-type II extracellular PLA2 antibodies},
  Life Sciences 61, PL127--PL134, (1997).

\bibitem{smith24}
  Smith, M. A. et al.,
  \textit{Treatment with dextroamphetamine decreases the reacquisition of cocaine self-administration: Consistency across social contexts},
  Drug and Alcohol Dependence 260, 111328, (2024).

\bibitem{chang14}
  Chang, Z. et al.,
  \textit{Stimulant ADHD medication and risk for substance abuse},
  Journal of Child Psychology and Psychiatry 55, 878--885, (2014).

\bibitem{mariani07}
  Mariani, J. J. \& Levin, F. R.,
  \textit{Treatment strategies for co-occurring ADHD and substance use disorders},
  The American Journal on Addictions 16, 45--56, (2007).

\bibitem{musco19}
  Musco, S. et al.,
  \textit{Characteristics of patients experiencing extrapyramidal symptoms or other movement disorders related to dopamine receptor blocking agent therapy},
  Journal of Clinical Psychopharmacology 39, 336--343, (2019).

\end{thebibliography}
\end{document}